\documentclass{article}
\usepackage{iclr2026_conference,times}

\usepackage{amsmath,amsfonts,bm}

\def\eqref#1{equation~\ref{#1}}

\def\Algref#1{Algorithm~\ref{#1}}

\def\1{\bm{1}}

\DeclareMathAlphabet{\mathsfit}{\encodingdefault}{\sfdefault}{m}{sl}
\SetMathAlphabet{\mathsfit}{bold}{\encodingdefault}{\sfdefault}{bx}{n}

\usepackage{hyperref}
\usepackage{xurl}
\usepackage{booktabs}
\usepackage{etoolbox}
\usepackage{amsfonts}
\usepackage{nicefrac}
\usepackage{microtype}
\usepackage{booktabs}
\usepackage{graphicx}
\usepackage{xcolor}
\usepackage{colortbl}
\usepackage{amsmath}
\usepackage{graphicx}
\usepackage{wrapfig}
\usepackage{fvextra}
\usepackage{enumitem}
\usepackage{makecell}
\usepackage{caption}
\usepackage[most]{tcolorbox}
\usepackage{tabularx}
\usepackage{amsthm}
\usepackage{bm}
\usepackage{multirow}
\usepackage[ruled]{algorithm2e}
\tcbuselibrary{breakable}
\usepackage[capitalize,noabbrev]{cleveref}
\usepackage{titletoc}
\usepackage{tikz}
\usepackage[table]{xcolor}
\definecolor{opdblue}{RGB}{225,240,250}
\definecolor{deltagreen}{RGB}{0,130,70}
\definecolor{groupgray}{RGB}{235,235,235}
\usepackage{tabularx}
\usepackage{array}
\usepackage{fontawesome5}

\tcbuselibrary{skins}

\definecolor{abstractbg}{rgb}{0.92, 0.96, 1}
\definecolor{abstractborder}{named}{iclrdeepblue}

\renewenvironment{abstract}{%
  \vskip 0.075in%
  \begin{tcolorbox}[
    colback=abstractbg,
    colframe=abstractborder,
    boxrule=1.2pt,
    arc=5pt,
    left=5pt, right=5pt, top=8pt, bottom=8pt,   
    boxsep=0pt,
    width=\textwidth,
    before skip=0pt, after skip=0pt
  ]%
  \centerline{\large\sc Abstract}%
  \vspace{0.5ex}%
  \begin{quote}%
}{%
  \end{quote}%
  \end{tcolorbox}%
  \vskip 1ex%
}

\newcommand{\deltaup}[1]{%
  \textcolor{deltagreen}{\scriptsize $(+\Delta\,#1)$}%
}

\definecolor{mintblue}{RGB}{210,235,250}
\definecolor{mintframe}{RGB}{120,180,220} 
\definecolor{minttitle}{RGB}{100,150,200} 
\definecolor{minttext}{RGB}{50,80,120}    

\definecolor{runzhemilk}{RGB}{255,235,245} 
\definecolor{roseframe}{RGB}{230,120,150}  
\definecolor{runzhecotton}{RGB}{255,170,200}    

\newtcolorbox{promptbox}[1]{
  enhanced,
  breakable,
  colback= runzhemilk!30!white,   
  colframe=roseframe,                
  colbacktitle= runzhecotton!66!white, 
  coltitle=white!33,
  title=\textbf{#1},
  fonttitle=\bfseries,
  sharp corners=south, 
  borderline={0.8pt}{0pt}{roseframe},
  boxrule=0.8pt,
  arc=6pt, 
  left=6pt, right=6pt, top=6pt, bottom=6pt,
  before skip=10pt, after skip=10pt,
  drop shadow=black!12,      
}

\definecolor{proofsolutionbg}{RGB}{248,247,252}
\definecolor{proofsolutionframe}{RGB}{156,143,181}
\definecolor{proofsolutiontitle}{RGB}{121,101,153}

\newtcolorbox{proofsolutionbox}[1]{
  enhanced,
  breakable,
  colback=proofsolutionbg,
  colframe=proofsolutionframe,
  colbacktitle=proofsolutiontitle,
  coltitle=white,
  title=\textbf{#1},
  fonttitle=\bfseries,
  sharp corners=south,
  borderline={0.8pt}{0pt}{proofsolutionframe},
  boxrule=0.8pt,
  arc=6pt,
  left=6pt, right=6pt, top=6pt, bottom=6pt,
  before skip=10pt, after skip=10pt,
  drop shadow=black!12,
}

\newtcolorbox{casebox}[1]{
enhanced,
breakable,
colback=mintblue!40!white,
colframe=mintframe,
colbacktitle=minttitle!70!white,
coltitle=white,
title=\textbf{#1},
fonttitle=\bfseries,
sharp corners=south, 
borderline={0.8pt}{0pt}{minttitle},
boxrule=0.8pt,
arc=6pt, 
left=6pt, right=6pt, top=6pt, bottom=6pt,
before skip=10pt, after skip=10pt,
drop shadow=black!15, 
}

\newtcolorbox{takeawaysbox}{
enhanced,
breakable,
colback=mintblue!40!white,
colframe=mintframe,
colbacktitle=minttitle!70!white,
coltitle=white,
title=\textbf{Key Takeaways},
fonttitle=\bfseries,
sharp corners=south, 
borderline={0.8pt}{0pt}{minttitle},
boxrule=0.8pt,
arc=6pt, 
left=6pt, right=6pt, top=6pt, bottom=6pt,
before skip=10pt, after skip=10pt,
drop shadow=black!15, 
}

\makeatletter
\newcommand{\DrawLine}{%
  \begin{tikzpicture}
  \path[use as bounding box] (0,0) -- (\linewidth,0);
  \draw[color=minttitle!70!white,dashed,dash phase=1.5pt]
        (0-\kvtcb@leftlower-\kvtcb@boxsep,0)--
        (\linewidth+\kvtcb@rightlower+\kvtcb@boxsep,0);
  \end{tikzpicture}%
  }
\makeatother

\definecolor{darkblue}{rgb}{0, 0, 0.5}
\definecolor{citeblue}{rgb}{0.35, 0.62, 0.88}
\hypersetup{colorlinks=true, citecolor=citeblue, urlcolor=citeblue, linkcolor=darkblue}

\newcommand{\projectpageurl}{}
\newcommand{\huggingfaceurl}{https://huggingface.co/collections/bingyang-lei/simpleopd}
\newcommand{\githuburl}{https://github.com/hhnqqq/SimpleOPD}
\newcommand{\projecturl}{https://hhnqqq.github.io/SimpleOPD-project-page}
\newcommand{\resourceIcon}[1]{\raisebox{-0.24em}{\includegraphics[height=1.05em]{#1}}}
\newcommand{\resourceLabel}[2]{\resourceIcon{#1}\hspace{0.35em}\textcolor{citeblue}{\textbf{#2}}}

\newcommand{\resourceLink}[3]{\href{#1}{\resourceLabel{#2}{#3}}}

\title{SimpleOPD: Simple Tokenizer-Agnostic On-Policy Distillation for Long-Context Reasoning}

\author{%
\begin{minipage}{0.96\textwidth}
\vspace*{0.3em}
\centering
\normalfont
{\small\textbf{Haonan He}\textsuperscript{*} \enspace
\textbf{Haodi Lei}\textsuperscript{*}\enspace
\textbf{Yun Luo}\textsuperscript{\textdagger,$\ddagger$}\enspace
\textbf{Haoran Zhang}\textsuperscript{}\enspace
\textbf{Shunkai Zhang}\textsuperscript{}\enspace
\textbf{Yizhuo Li}\textsuperscript{}\enspace \\
\textbf{Shengji Tang}\textsuperscript{}\enspace 
\textbf{Zhilin Wang}\textsuperscript{}\enspace 
\textbf{Runzhe Zhan}\textsuperscript{}\enspace 
\textbf{Lei Bai}\textsuperscript{}\enspace 
\textbf{Ganqu Cui}\textsuperscript{}\enspace \\
\textbf{Fangchen Yu}\textsuperscript{$\ddagger$}\enspace  
\textbf{Yafu Li}\textsuperscript{$\ddagger$}\enspace
\textbf{Peng Ye}\textsuperscript{$\ddagger$}\enspace
\textbf{Ning Ding}\textsuperscript{$\ddagger$}\enspace
\textbf{Yu Cheng}\textsuperscript{$\ddagger$}
}\\[0.7em]
{\footnotesize
SU-01 Team, Shanghai Artificial Intelligence Laboratory  \\
\textsuperscript{*}Equal Contribution,$\quad$\textsuperscript{\textdagger}Project Lead,$\quad$
\textsuperscript{$\ddagger$}Corresponding Author \\
\faEnvelope\hspace{0.35em}\texttt{luoyun@pjlab.org.cn}
}
\end{minipage}%
}

\iclrfinalcopy

\begin{document}

\maketitle

\begingroup
\renewcommand{\thefootnote}{\fnsymbol{footnote}}
\footnotetext[1]{Work done during an internship at Shanghai Artificial Intelligence Laboratory.}
\endgroup
\vspace{-5mm}

\begin{abstract}
On-policy distillation (OPD) offers a promising way to transfer reasoning capabilities from stronger teacher models, but applying it to long-context reasoning teachers and short-context students introduces practical challenges, including tokenizer mismatch, teacher-student distribution mismatch, response length explosion, and training instability. In this work, we study this setting by transferring proof-reasoning capabilities from the long-context reasoning model SU-01 to short-context student models. To handle tokenizer differences, we perform OPD in a shared text space and align only tokens that occupy identical text spans under the student and teacher tokenizers. 
To mitigate the problem of excessive generation length and frequent truncation, we introduce a student reference KL loss and mask the advantages of special termination tokens such as ${\scriptstyle <}\,\mathrm{/think}{\scriptstyle >}$ and ${\scriptstyle <}\,\mathrm{|im\_end|}\,{\scriptstyle >}$. This strategy constrains the student from drifting excessively from its initial policy, thereby mitigating the teacher-student distribution mismatch problem and fostering steady length growth. Experiments on both same-family and different-family student models, including Qwen3, Qwen3.5, Intern-S2, GLM-4.7, Gemma-4, show consistent gains in mathematical reasoning, especially natural-language math proving. Notably, Intern-S2-Preview improves by 21.2 points on ProofBench, reaching 55.2 and surpassing Gemini-2.5-Pro. It also improves on science benchmarks such as HLE and HiPhO, suggesting that OPD transfers reasoning capabilities that generalize beyond the mathematical training domain.

\vspace{1mm}

\begin{center}
{\normalsize
\resourceLink{\projecturl}{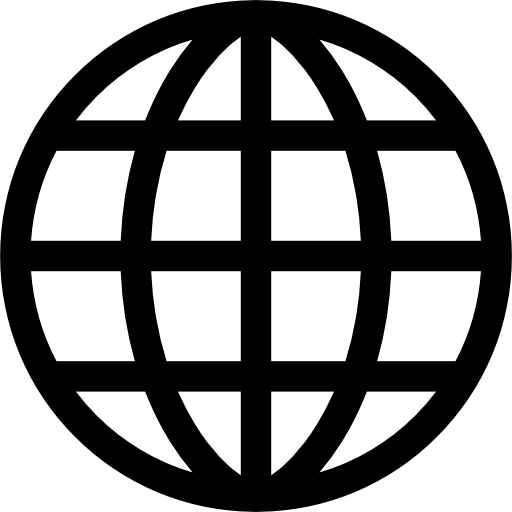}{Project Page}
\hspace{2em}
\resourceLink{\githuburl}{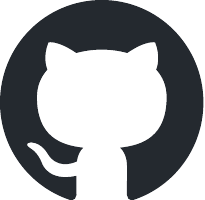}{Code}
\hspace{2em}
\resourceLink{\huggingfaceurl}{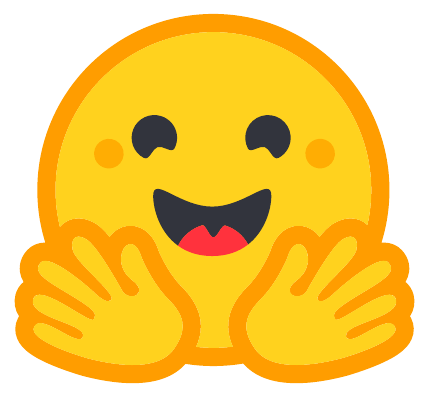}{Models}
\hspace{2em}
\textcolor{gray}{\faCalendar}\hspace{0.3em}August, 2026
}
\end{center}
\end{abstract}

\begin{figure}[hp]
    \centering
    \includegraphics[width=\linewidth]{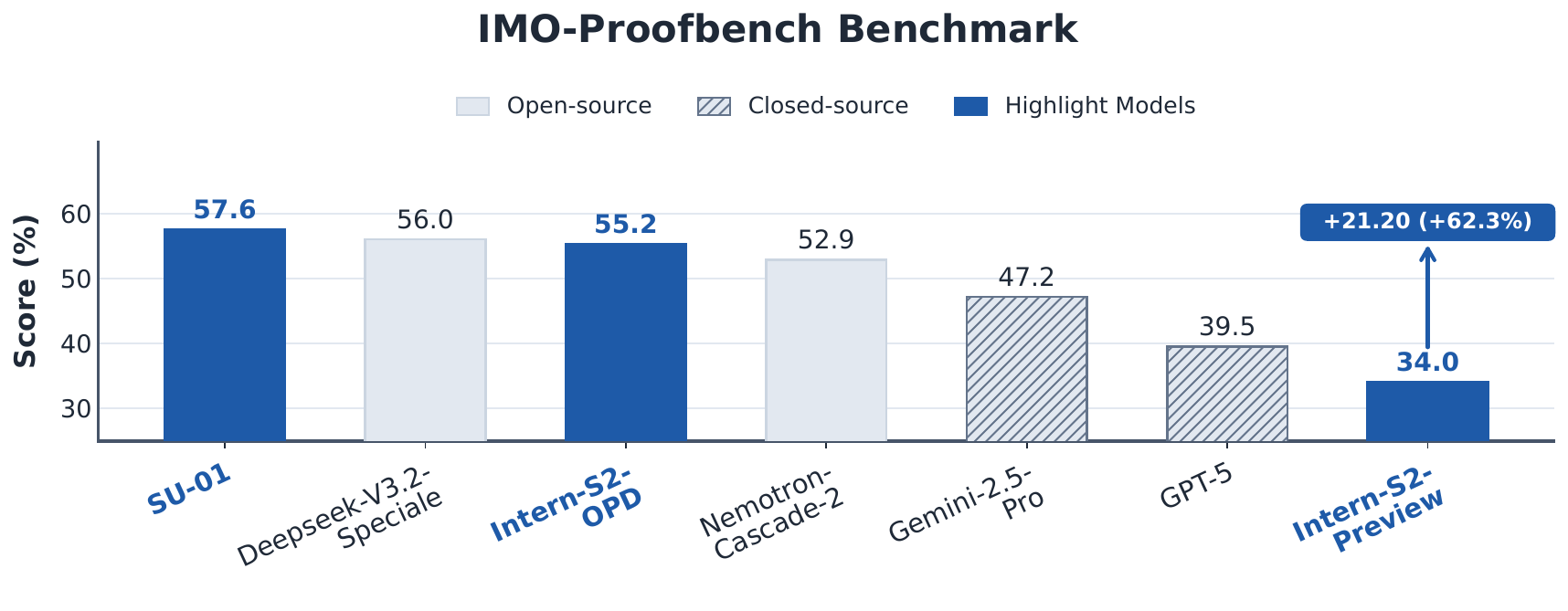}
    \vspace{-3mm}
    \caption{Evaluation performance on ProofBench using Gemini-2.5-Pro as the judge, under the same evaluation setting as SU-01. With SimpleOPD, Intern-S2-OPD achieves performance comparable to DeepSeek-V3.2-Speciale, demonstrating the effectiveness of our OPD strategy.}
    \label{fig:gemini2.5}
\end{figure}

\newpage
\setcounter{tocdepth}{2}
\renewcommand{\contentsname}{Contents}
\tableofcontents
\newpage

\section{Introduction}
\label{sec:introduction}
On-policy distillation (OPD)~\citep{gu2024minillm,agarwal2024onpolicy} presents a compelling paradigm in the post-training landscape of LLMs for tasks such as strong-to-weak distillation~\citep{lu2025onpolicydistillation}, multi-teacher distillation~\citep{ma2026mopdmultiteacheronpolicydistillation}, and self-distillation~\citep{hübotter2026reinforcementlearningselfdistillation}. Unlike off-policy approaches that distill knowledge from fixed teacher-generated data, OPD actively queries the teacher to evaluate tokens in student-generated trajectories, thereby providing dense token-level supervision grounded in the student's own policy \citep{gu2024minillm,agarwal2024onpolicy}, 
enabling efficient distillation of complex reasoning behaviors, as it incurs less forgetting and achieves better generalization than off-policy settings. 
However, prior work has primarily studied teacher–student pairs that share the same model family and vocabulary. Distilling reasoning capabilities from long-context teachers into short-context students, especially across different model families, remains largely underexplored~\citep{sun2026simct}.

Applying OPD to distilling long-context reasoning teachers to short-context reasoning students presents several challenges. First, teacher and student models may use different tokenizers, making direct alignment between their token-level distributions infeasible~\citep{zhang-etal-2024-dual,chen-etal-2025-enhancing-cross,singh-etal-2026-cross}. Second, differences in model capacity and context length introduce a substantial teacher–student distribution mismatch~\citep{li2026rethinkingopd}. In particular, long-context reasoning teachers often favor responses that exceed the student’s context budget. Through experiments, we observe that direct OPD can cause rapid response-length growth, frequent truncation, and incomplete reasoning trajectories, ultimately destabilizing training. This issue is especially pronounced for termination tokens such as ${\scriptstyle <}\,\mathrm{/think}{\scriptstyle >}$ and ${\scriptstyle <}\,\mathrm{|im\_end|}\,{\scriptstyle >}$, as teacher supervision may repeatedly discourage the student from terminating its thinking process or final response.

In this work, we focus on the transfer of reasoning capabilities from the long-context reasoning model SU-01 \citep{li2026su01} to short-context student models. As shown in Figure~\ref{fig:simpleopd}, to enable distillation across different model families, we perform OPD in a shared text space and align teacher and student tokens only when they occupy identical text spans. This provides reliable token-level supervision without requiring an artificial correspondence between incompatible tokenizations. To mitigate excessive response-length growth and stabilize optimization, we introduce a reference KL-divergence loss between the student and its initial policy. We also mask the advantages of special termination tokens, preventing teacher supervision from directly suppressing termination. Together, these techniques constrain excessive policy drift, alleviate teacher--student distribution mismatch, and promote steady response-length growth within the student's context budget.

We evaluate SimpleOPD across both same-family and cross-family distillation settings, using student models from Qwen3, Qwen3.5, Intern-S2-Preview, GLM-4.7, and Gemma-4 \citep{yang2025qwen3,qwenteam2026qwen35,internlm2026interns2preview,zai2026glm47flash,gemmateam2026gemma4}. Across model families, SimpleOPD consistently improves mathematical reasoning performance, with particularly strong gains on natural-language mathematical proof tasks. Most notably, Intern-S2-Preview improves by 21.2 points on ProofBench~\citep{luong-etal-2025-towards}, from 34.0 to 55.2, surpassing Gemini-2.5-Pro and approaching the performance of substantially stronger reasoning models (Figure~\ref{fig:gemini2.5}). The distilled students also achieve improvements on science-oriented benchmarks, including HLE and HiPhO~\citep{phan2025humanity,yu2025hipho}, despite being trained exclusively on mathematical reasoning data. 
These results demonstrate that stable cross-tokenizer OPD can transfer reasoning capabilities from long-context teachers, and that the acquired capabilities generalize beyond the training domain.

In summary, our observations and contributions are:

\begin{figure*}[t]
\vspace{-3mm}
    \centering
    \includegraphics[width=\textwidth]{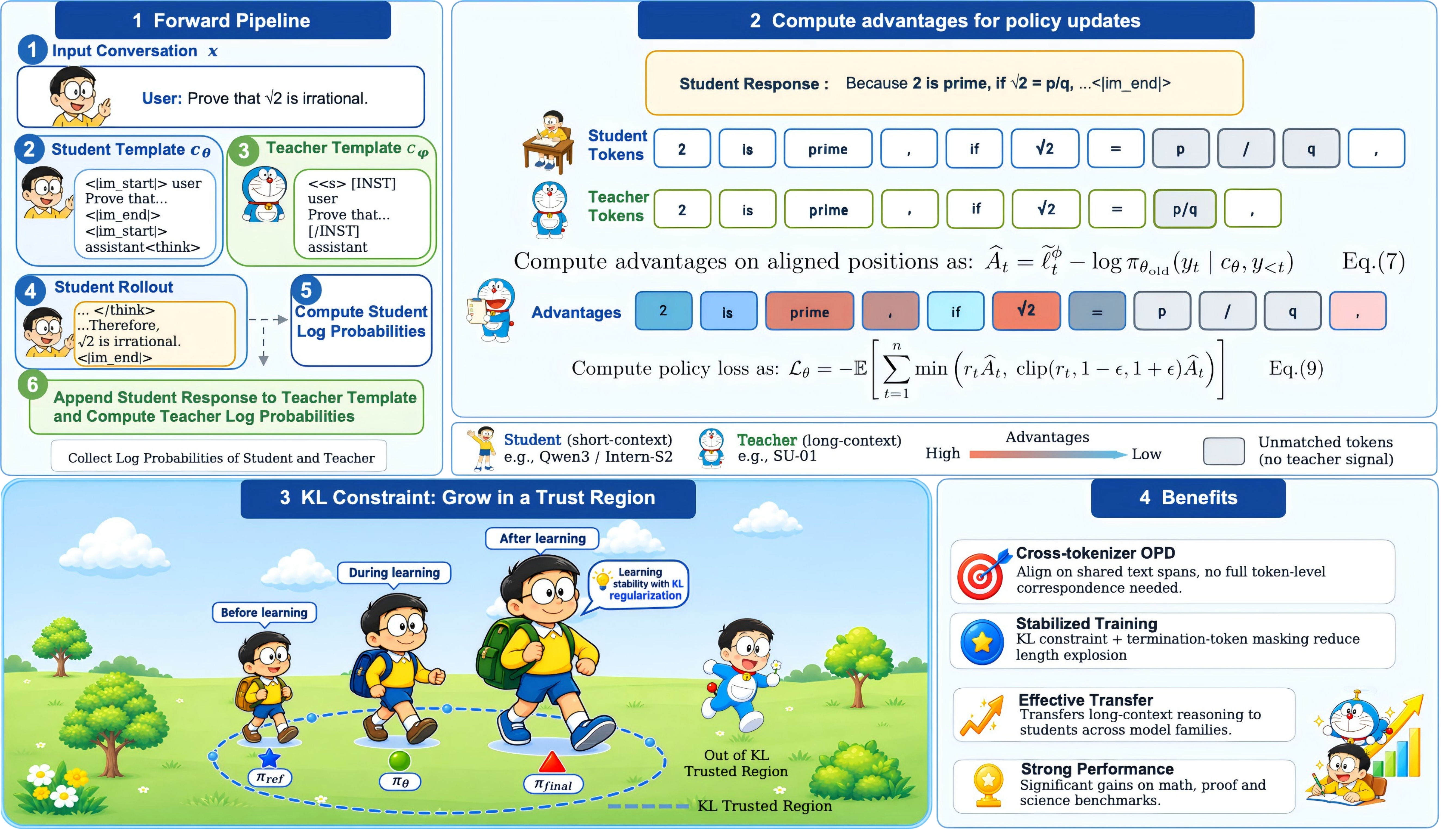}
    \caption{
    Overview of SimpleOPD. 
    The student generates responses under its own tokenizer, while the teacher evaluates the same response under its native tokenizer. 
    We align tokens with identical text spans, use teacher supervision for matched tokens, and stabilize training with KL regularization and termination-token masking.
    }
    \label{fig:simpleopd}
\end{figure*}

\begin{itemize}
\item Cross-tokenizer OPD can improve student models without full token-level alignment by using student-generated text as a shared space and aligning tokens with identical text spans.
\item Direct OPD transfers long-context reasoning capabilities, but it often causes excessive length growth, frequent truncation, and training instability.
\item Termination-token advantage masking and student reference KL loss stabilize training and reduce truncation. Larger teacher--student gaps generally require stronger KL regularization.
\item SimpleOPD shows consistent performance gains across model families without supervised fine-tuning on teacher trajectories, while moderately increasing the distillation length can further improve long-form reasoning.
\end{itemize}

\section{Method}
\label{sec:method}

Let $\mathbf{x}$ denote the input conversation represented as a list of messages. The student and teacher with different tokenizers may use different chat templates, denoted by $\mathcal C_\theta$ and $\mathcal C_\phi$, respectively. We first construct the student input context text $c_\theta=\mathcal C_\theta(\mathbf{x})$ and sample a response-token sequence with the student policy executed by the rollout engine $\pi_{\theta_{\mathrm{roll}}}$:
\begin{equation}
   y_{1:n}=(y_1,\ldots,y_n)
\sim
\pi_{\theta_{\mathrm{roll}}}(\cdot\mid c_\theta). 
\end{equation}

Decoding $y_{1:n}$ via the tokenizer decoder of the student gives the response surface string $s=\mathcal D_\theta(y_{1:n})$. Instead of passing the student context to the teacher, we reconstruct the teacher context text $c_\phi$ using its own chat template and append the student response:
$$
c_\phi=\mathcal C_\phi(\mathbf{x}),
\quad
u_\phi=c_\phi\oplus s,
$$
where $\oplus$ denotes string concatenation. The complete text $u_\phi$ is then provided to the teacher, which tokenizes it using the tokenizer encoder $\mathcal E_\phi$of the teacher. Let $\mathcal E_\phi(s|c_\phi)=(z_1,\ldots,z_m)$ denote token list obtained by encoding the student response given the input context $c_\phi$using the teacher tokenizer. This procedure allows the teacher to evaluate the student response under its native tokenizer and chat-template. Although $c_\theta$ and $c_\phi$ may differ, the response text being evaluated is identical on both sides.
For token alignment, let $\tau_\theta(y_t)$ and $\tau_\phi(z_i)$ denote the incremental text spans contributed by the corresponding response tokens. They satisfy
$$
\bigoplus_{t=1}^{n}\tau_\theta(y_t)
=
\bigoplus_{i=1}^{m}\tau_\phi(z_i)
=
s.
$$
We perform no additional cleanup or normalization, so these spans define consistent offsets in the shared response string. The teacher log-probability associated with token $z_i$ is
$\log\pi_\phi(z_i\mid c_\phi,z_{<i})$.

\subsection{Cross-tokenizer Alignment}
Define the cumulative response text preceding each student and teacher token as
$$
P_\theta(t)=\bigoplus_{k=1}^{t-1}\tau_\theta(y_k),\qquad P_\phi(i)=\bigoplus_{k=1}^{i-1}\tau_\phi(z_k),
$$
with $P_\theta(1)=P_\phi(1)=\varepsilon$, where $\varepsilon$ denotes the empty string. A teacher position $i$ is aligned with a student position $t$ if both tokenizations have consumed the same response prefix and the current tokens contribute the same text span:
\begin{equation}
    \mathcal M=\{(i,t):P_\phi(i)=P_\theta(t)\land\tau_\phi(z_i)=\tau_\theta(y_t)\}.
\end{equation}
Equivalently, an aligned teacher-student pair covers the same start and end offsets in the shared response string $s$. Tokens that overlap only partially are not aligned because the log‑probability of one teacher token cannot be uniquely assigned to multiple student tokens, nor can the log‑probabilities of multiple teacher tokens be uniquely merged into a single student token. Since $y$ and $z$ are ordered segmentations of the same response string, $\mathcal M$ is a partial one-to-one mapping: each token on either side can match at most one token on the other side.
We enumerate $\mathcal M$ using a linear two-pointer scan. The scan maintains the accumulated response prefixes on both sides. If the prefixes and current token spans agree, the two tokens are aligned and both pointers advance. Otherwise, the side that has consumed less response text advances until the two prefixes meet again. When both sides have consumed the same amount of text but the current token spans differ, both pointers advance.


For each student position $t$, define the alignment indicator
\begin{equation}
    a_t=\mathbf{1}[\exists i \text{ such that } (i,t)\in\mathcal M].
\end{equation}
Let $\log \pi_\theta (y_t \mid c_{\theta}, y_{<t})$ and $\log \pi_{\phi} (z_{i} \mid c_{\phi}, z_{<{i}})$ denote the log-probabilities of student policy and teacher policy, respectively. We construct a student-length teacher target as
\begin{equation}
    \widetilde{\ell}_t^\phi=
  \begin{cases}
  \log \pi_{\phi} (z_{i} \mid c_{\phi}, z_{<{i}}), & a_t=1,\\
  \log \pi_\theta (y_t \mid c_{\theta}, y_{<t}), & a_t=0.
\end{cases}
\end{equation}
Thus, aligned positions inherit the corresponding teacher log-probability, while unmatched positions fall back to the student's log-probability. We report the lexical overlap ratio
$
\rho=\frac{|\mathcal M|}{n},
$
which measures the fraction of student response tokens that receive teacher supervision. The complete alignment procedure is provided in \Algref{alg:cross-tokenizer-alignment} in Appendix~\ref{app:cross-tokenizer-alignment}.

\subsection{On-policy distillation objective}

The cross-tokenizer distillation objective is defined over aligned response positions:
\begin{equation}
    \mathcal L_{\mathrm{Distill}}(\theta)=\mathbb E_{y\sim\pi_\theta}\left[\sum_{t=1}^{n}\log\pi_\theta(y_t\mid c_\theta,y_{<t})-\widetilde{\ell}_{t}^\phi\right].
\end{equation}
This objective is a token-aligned surrogate for the reverse KL divergence. It compares the teacher and student probabilities only at positions where the two tokenizers induce the same local segmentation of the response string.
When the tokenizers are identical, every student token is aligned with the corresponding teacher token, so $a_t=1$ and $z_t=y_t$ for all $t$. The objective then reduces to
\begin{equation}
    \mathcal L_{\mathrm{Distill}}(\theta)=\mathbb E_{y\sim\pi_\theta}\left[\log\frac{\pi_\theta(y\mid c_\theta)}{\pi_\phi(y\mid c_\phi)}\right]=D_{\mathrm{KL}}\left(\pi_\theta(\cdot\mid c_\theta)\parallel
\pi_\phi(\cdot\mid c_\phi)\right).
\end{equation}
When the tokenizers differ, the same response string is factorized into different token sequences, so the exact token-level KL is not directly computable. The proposed objective instead applies teacher supervision only where the two tokenizations agree on both the token boundary and textual span.
To enable multiple policy updates on the same rollout batch, we substitute the online log-probability $\log \pi_\theta (y_t \mid c_{\theta}, y_{<t})$ in $\widetilde{\ell}_t^\phi$ with its pre-update counterpart $\log \pi_{\theta_\text{old}}(y_t \mid c_{\theta}, y_{<t})$ for all unmatched positions ($a_t = 0$) and define the fixed policy advantages as
\begin{equation}
  \widehat A_t = \widetilde{\ell}_t^\phi-\log \pi_{\theta_\text{old}} (y_t \mid c_{\theta}, y_{<t}),  
\end{equation}
using the PPO clipped policy loss. The importance-sampling ratio is
\begin{equation}
  r_t
=
\frac{
\pi_\theta(y_t\mid c_\theta,y_{<t})
}{
\pi_{\theta_{\mathrm{old}}}(y_t\mid c_\theta,y_{<t})
}.  
\end{equation}

The resulting objective is
\begin{equation}
  \mathcal L_\theta
=
-\mathbb E
\Bigg[
\sum_{t=1}^{n}
\min\Big(
r_t\widehat A_t,\;
\operatorname{clip}(r_t,1-\epsilon,1+\epsilon)\widehat A_t
\Big)
\Bigg].  
\end{equation}

\section{Experiment}
\label{sec:experiment}
We adopt SU-01, an IMO gold-medal-level mathematical reasoning model developed by our team, as the teacher model. Our goal is to transfer its mathematical reasoning and proof-generation capabilities to a diverse set of student models through on-policy distillation.

We consider both same-vocabulary and cross-tokenizer settings. For the same-tokenizer setting, we distill SU-01 into models from the Qwen3 family, including Qwen3-4B-thinking-2507 and Qwen3-30B-A3B-Thinking-2507 (abbreviated as Qwen3-4B and Qwen3-30B-A3B, respectively). For the cross-tokenizer setting, we further evaluate models with different tokenizer designs, including Qwen3.5-4B, Qwen3.5-35B-A3B, Intern-S2-Preview, GLM-4.7-Flash, and Gemma-4-26B-A4B-it (abbreviated as Gemma-4-26B-A4B). This setting allows us to examine whether cross-tokenizer OPD can transfer mathematical proof capabilities across different model architectures and tokenization schemes.

\begin{figure}
    \centering
    \includegraphics[width=\linewidth]{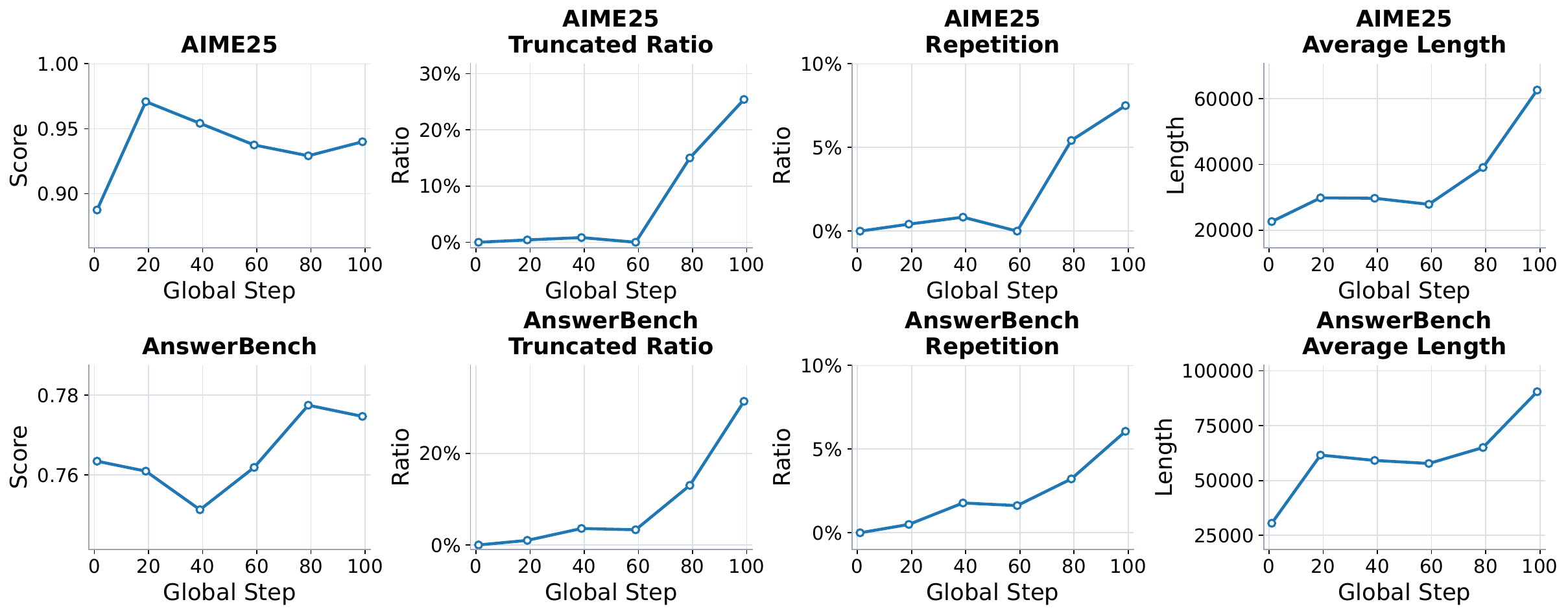}
    \caption{Training dynamics of Intern-S2-Preview during direct OPD training.}
    \label{fig:vallina}
\end{figure}

\subsection{Training Data} 

All training examples are mathematical proof problems. The collection combines curated proof corpora, community-contributed problems, and olympiad training materials to provide a varied set of proof-oriented prompts. Our training set consists of:
\begin{enumerate}
    \item \textbf{OPC}: 63 problems from the Open Proof Corpus.
    \item \textbf{AoPS}: 2,948 problems collected from the Art of Problem Solving community.
    \item \textbf{Books}: 900 problems collected from online mathematical competition training books.
    \item \textbf{Shuzhimi}: 617 problems sourced from the Shuzhimi Forum, an online Chinese mathematical problem-solving community, and Evan Chen’s olympiad materials.
\end{enumerate}

\subsection{Implementation Details}

We perform OPD using responses sampled from the student policy, with \textbf{SU-01} serving as the teacher model. SU-01 is a 30B-A3B reasoning model based on Qwen3-30B-A3B with strong mathematical proof capabilities, achieving gold-medal-level performance on olympiad-level evaluations and demonstrating particularly competitive results on proof-oriented benchmarks such as ProofBench. It is also capable of sustaining over 100K tokens of natural-language reasoning for difficult Olympiad problems, making it a powerful yet highly long-context teacher. During distillation, SU-01 provides token-level supervision at positions that can be aligned across the teacher and student tokenizations.

\begin{figure}
    \centering
    \includegraphics[width=\linewidth]{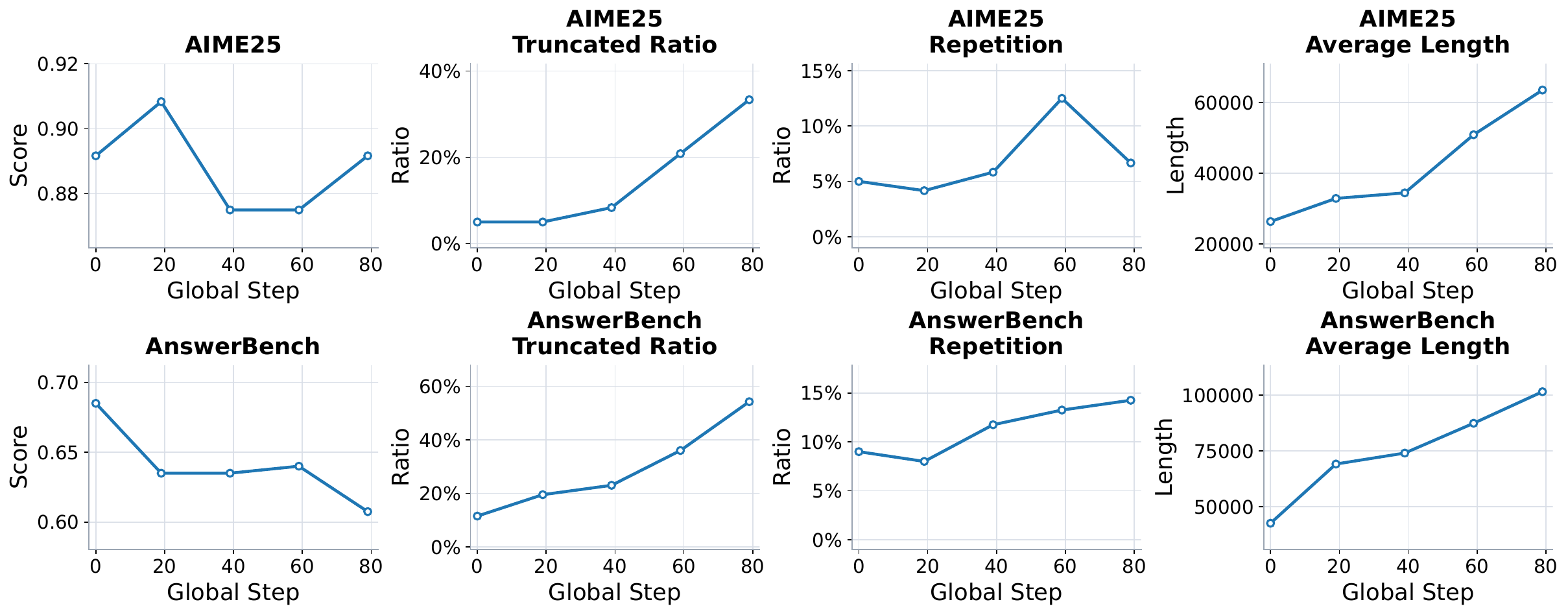}
    \caption{Training dynamics of Qwen3.5-35B-A3B during direct OPD training.}
    \label{fig:qwen35vallina}
\end{figure}

We use Slime~\citep{slime_github} as the underlying framework for OPD and SGLang~\citep{zheng2024sglang} for evaluation rollout. The models are trained for 100 rollout iterations with a constant learning rate of 1e-6, a rollout batch size of 64, 4 responses sampled per prompt, and a maximum rollout length of 32K tokens for Qwen-Series, 6k for GLM-4.7 and Gemma-4.
The policy objective uses a PPO clipping coefficient of 0.2. The policy is updated four times per rollout step.

In our evaluation, we primarily evaluate the non-verifiable ProofBench and the verifiable AnswerBench from the IMO-Bench suite~\citep{luong-etal-2025-towards}, together with the verifiable benchmarks AIME25~\citep{aime2025} and AMOBench~\citep{an2025amobench}. For ProofBench, we use DeepSeek-V4-Flash as the judge for cost-efficient evaluation. To mitigate evaluation instability, we evaluate 4 times for the same proof rollout and choose the average result. For the latter three, we first use a rule-based verifier for testing; if the answer is correct, it is counted as correct, and if it is wrong, we then use GPT-OSS-120B~\citep{openai2025gptoss120bgptoss20bmodel} for evaluation. The reported results are the averages over 4 rollouts for ProofBench and 8 for the other three.

We use the following evaluation settings: temperature 1.0, top-p 0.95, repetition penalty 1.0, and a maximum response length of 160,000 tokens.  The best checkpoint is selected based on the average score of AIME@4 and AnswerBench@1.

\section{Experimental Results}
\label{sec:analysis}
\subsection{Training Instability in On-Policy Distillation}
In this section, we first examine training instability in direct on-policy distillation, including length explosion and truncation. Accordingly, we propose two mitigation strategies, namely masking special tokens and adding a student reference KL loss. 

\subsubsection{Initial Observation: Length Explosion and Truncation}
During OPD training, the student model Intern-S2-Preview shows modest improvements in task performance, but these gains are accompanied by clear signs of degeneration (Figure \ref{fig:vallina}). Specifically, the truncation rate and repetition rate both increase substantially as training progresses, while the average response length also grows sharply. This phenomenon is even more pronounced for Qwen3.5-35B-A3B (Figure \ref{fig:qwen35vallina}), as its task performance deteriorates more noticeably, and the degradation is accompanied by a persistently high truncation rate and rapidly increasing response length. These results suggest that apparent performance gains can come at the cost of increasingly verbose and unstable generation behavior, revealing a trade-off between task performance and output quality. Case studies could be found in Appendix \ref{app:case}.

\subsubsection{Special Token Masking}
During training, we observe that the student model’s output length keeps increasing, and many generated responses fail to emit termination tokens such as ${\scriptstyle <}\,\mathrm{|im\_end|}\,{\scriptstyle >}$. This suggests that the student is strongly affected by the long outputs from the teacher and gradually loses the ability to terminate properly. To mitigate this issue, we mask the OPD loss on the structural tokens ${\scriptstyle <}\,\mathrm{/think}{\scriptstyle >}$ and ${\scriptstyle <}\,\mathrm{|im\_end|}\,{\scriptstyle >}$, as these tokens primarily control the output format and termination behavior and do not need to match the teacher distribution exactly. 

\begin{figure}
    \centering
    \includegraphics[width=\linewidth]{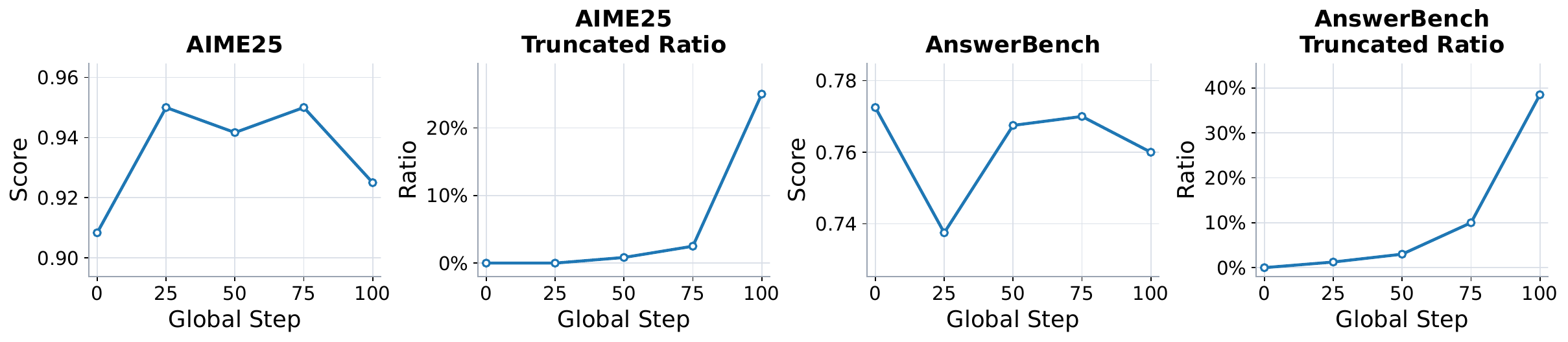}
    \caption{Intern-S2-Preview OPD training dynamics when masking the special tokens such as ${\scriptstyle <}\,\mathrm{/think}{\scriptstyle >}$ and ${\scriptstyle <}\,\mathrm{|im\_end|}\,{\scriptstyle >}$. The special-token masking alone helps mitigate length-related instability but cannot resolve the length expansion problem.}
    \label{fig:specialmask}
\end{figure}

The validation curve is shown in Figure \ref{fig:specialmask}.         
The results suggest that the masking strategy helps mitigate length-related instability. However, in the later stage of training, the model’s length truncation rate still continues to increase sharply, indicating that special-token masking alone cannot fully resolve the length expansion problem. The evaluation results are shown in Table \ref{tab:spec_mask}. The results show that masking the advantages of special termination tokens improves OPD training stability, but special token masking alone cannot fully resolve the length-expansion problem.

\begin{figure}
    \centering
    \includegraphics[width=\linewidth]{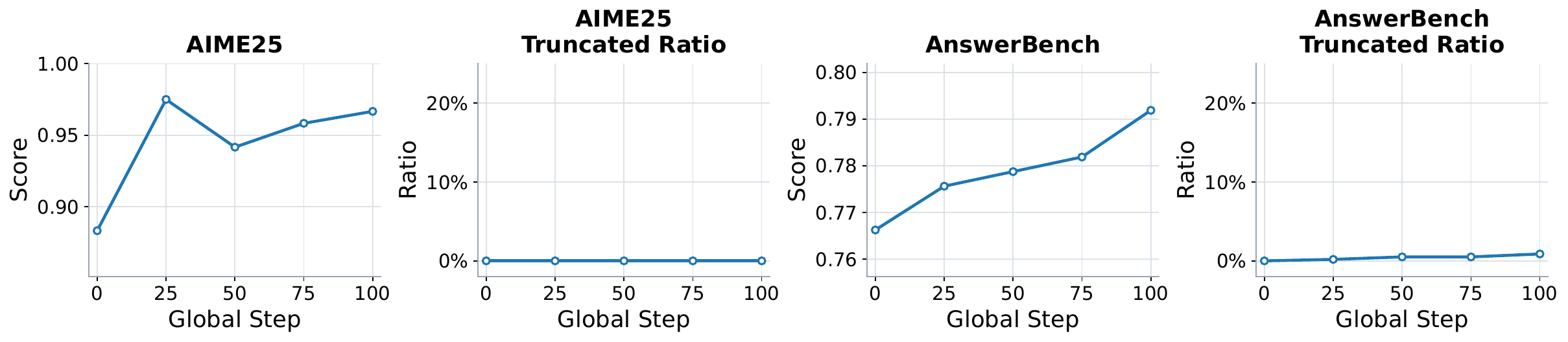}
    \caption{Intern-S2-Preview OPD training dynamics when adding student reference KL loss to prevent the student policy from deviating excessively from its initial distribution.  The training could be stabilized, and the truncation rate is effectively reduced to nearly zero.}
    \label{fig:sturef}
\end{figure}

\begin{table}[t]
    \centering
    \caption{OPD results from SU-01 to Intern-S2-Preview with masking termination tokens and adding student reference KL loss.}
    \label{tab:spec_mask}
    \small
    \renewcommand{\arraystretch}{1.1}
    \begin{tabular}{lccc}
        \toprule
        Model & ProofBench@4 & AnswerBench@8 & AIME25@8 \\
        \midrule
        SU-01 & 45.00 & 77.50 & 94.60 \\
        Intern-S2-Preview & 21.70 & 76.03 & 88.33 \\
        \rowcolor{opdblue}
        OPD + Spec Mask
            & 38.10 \deltaup{16.40}
            & 77.60 \deltaup{1.57}
            & 95.00 \deltaup{6.67} \\
                \rowcolor{opdblue}
        OPD + Ref KL
            & 38.50 \deltaup{16.80}
            & 79.10 \deltaup{3.07}
            & 95.80 \deltaup{7.47} \\
        \bottomrule
    \end{tabular}
\end{table}

\subsubsection{Student-Reference KL Loss}
We then attempt to introduce a student reference KL loss to prevent the student policy from deviating excessively from its initial distribution, thereby preserving the student model’s general capabilities during distillation. In this experiment, we set the student-reference kl-loss coefficient to 0.5. 

The validation curves during training are shown in Figure \ref{fig:sturef}.
As observed, the truncation rate is effectively reduced to nearly zero, while performance on AIME25 and AnswerBench is consistently improved. This indicates that the student reference KL loss can effectively mitigate the dramatic increase in response length during OPD training. The evaluation results are summarized in Table \ref{tab:spec_mask}.
The results demonstrate that adding the reference KL loss substantially improves OPD training while preserving the student model’s original capabilities. Specifically, OPD + Ref KL improves ProofBench@4 from 21.70 to 38.50, and further raises AnswerBench@8 and AIME25@8 to 79.10 and 95.80, respectively.


\subsection{Results on Same-family Models}
Finally, we combine special token masking with the student reference KL loss, which yields larger performance improvements. We set kl-loss-coef to 0.5 for the Qwen-series models and Intern-S2-Preview. 

\begin{table}[t]
    \centering
    \caption{Evaluation results of distilling SU-01 to students on ProofBench, AnswerBench, AIME25, and AMOBench.
    Green values indicate improvements over the corresponding base model.}
    \label{tab:main_results}
    \small
    \setlength{\tabcolsep}{4pt}
    \renewcommand{\arraystretch}{1.1}
    \begin{tabular}{lcccc}
        \toprule
        Model & ProofBench@4 & AnswerBench@8 & AIME25@8 & AMOBench@8 \\
        \midrule
        SU-01
            & 45.00 & 77.50 & 94.60 & 61.75 \\

        \rowcolor{groupgray}
        \midrule
        \multicolumn{5}{c}{Same-Tokenizer Models} \\
        \midrule

        Qwen3-4B
            & 11.42 & 47.50 & 71.25 & 23.00 \\
        \rowcolor{opdblue}
        Qwen3-4B-OPD
            & \textbf{23.72} \deltaup{12.30}
            & \textbf{64.50} \deltaup{17.00}
            & \textbf{90.83} \deltaup{19.58}
            & \textbf{35.00} \deltaup{12.00} \\

        Qwen3-30B-A3B
            & 13.80 & 59.13 & 88.33 & 36.50 \\
        \rowcolor{opdblue}
        Qwen3-30B-A3B-OPD
            & \textbf{36.47} \deltaup{22.67}
            & \textbf{74.46} \deltaup{15.33}
            & \textbf{93.75} \deltaup{5.42}
            & \textbf{52.75} \deltaup{16.25} \\

        \midrule
        \rowcolor{groupgray}
        \multicolumn{5}{c}{Cross-tokenizer Models} \\
        \midrule

        Qwen3.5-4B
            & 15.90 & 60.94 & 86.67 & 32.00 \\
        \rowcolor{opdblue}
        Qwen3.5-4B-OPD
            & \textbf{28.61} \deltaup{12.71}
            & \textbf{67.84} \deltaup{6.90}
            & \textbf{91.67} \deltaup{5.00}
            & \textbf{51.25} \deltaup{19.25} \\

        Qwen3.5-35B-A3B
            & 26.78 & 73.16 & 94.60 & 57.25 \\
        \rowcolor{opdblue}
        Qwen3.5-35B-A3B-OPD
            & \textbf{42.39} \deltaup{15.61}
            & \textbf{80.15} \deltaup{6.99}
            & \textbf{96.66} \deltaup{2.06}
            & \textbf{61.25} \deltaup{4.00} \\

        Intern-S2-Preview
            & 21.70 & 76.03 & 88.33 & 58.00 \\
        \rowcolor{opdblue}
        Intern-S2-OPD
            & \textbf{44.50} \deltaup{22.80}
            & \textbf{80.10} \deltaup{4.07}
            & \textbf{95.00} \deltaup{6.67}
            & \textbf{59.50} \deltaup{1.50} \\
        \bottomrule
    \end{tabular}
\end{table}

The experimental results are shown in Table \ref{tab:main_results}. Overall, SimpleOPD consistently improves performance across model sizes, tokenizer settings, and model families. Among the same-tokenizer models, Qwen3-4B-OPD gains 12.30 points on ProofBench, 17.00 points on AnswerBench, and 19.58 points on AIME25. Scaling the student to Qwen3-30B-A3B produces even greater proof-reasoning improvements; its ProofBench score increases by 22.67 points, from 13.80 to 36.47, while its AnswerBench and AIME25 scores improve by 15.33 and 5.42 points, respectively.

The improvements also generalize to cross-tokenizer distillation. Qwen3.5-35B-A3B-OPD reaches 42.39 on ProofBench, improving over its base model by 15.61 points, and achieves additional gains of 6.99 and 2.06 points on AnswerBench and AIME25. Intern-S2-Preview exhibits the largest ProofBench gain, improving by 22.80 points from 21.70 to 44.50, nearly matching the teacher’s score of 45.00. It also reaches 80.10 on AnswerBench and 95.00 on AIME25, surpassing SU-01 on both benchmarks. These results demonstrate that our approach effectively transfers proof-reasoning capabilities to diverse students without requiring shared tokenization or supervised fine-tuning on teacher-generated trajectories.


We also evaluate ProofBench using Gemini-2.5-Pro as the judge, following the same evaluation setting as SU-01. The results are shown in Figure \ref{fig:gemini2.5}.
The results show that OPD training brings a substantial improvement to Intern-S2-Preview on ProofBench. Specifically, Intern-S2-OPD improves from 34.0 to 55.2, achieving a gain of 21.2 points over the original Intern-S2-Preview model. This demonstrates that OPD can effectively transfer proof-reasoning capabilities from the long-context teacher model to the student model.
Notably, Intern-S2-OPD also surpasses several strong frontier models, including Gemini-2.5-Pro and GPT-5. Although it still trails SU-01 and DeepSeek-V3.2-Speciale, the gap is significantly narrowed after OPD training. These results further confirm the effectiveness of OPD for improving natural-language mathematical proof reasoning, especially in cross-model capability transfer settings.


\subsection{Results on Cross-family Models}
\begin{wrapfigure}{r}{0.35\textwidth}
    \vspace{-2.0em}
    \centering
    \captionsetup{font=small,skip=2pt}
    \includegraphics[width=\linewidth]{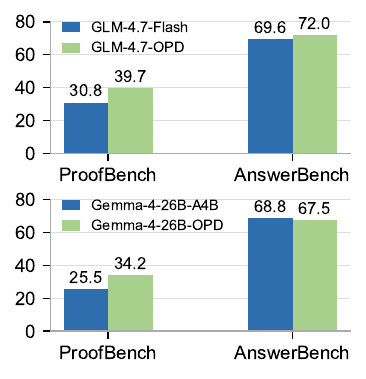}
    \caption{{Cross-family distillation results for GLM-4.7-Flash and Gemma-4-26B-A4B from SU-01.}}
    \label{fig:cross-family}
    \vspace{-3em}
\end{wrapfigure}

We set student-reference KL loss 1.0 for GLM-4.7-Flash and Gemma-4-26B-A4B to further maintain stability. Figure~\ref{fig:cross-family} presents additional cross-family distillation results on GLM-4.7-Flash and Gemma-4-26B-A4B. GLM-4.7-OPD improves consistently on both benchmarks, increasing from 30.8 to 39.7 on ProofBench and from 69.6 to 72.0 on AnswerBench. Gemma-4-26B-A4B-OPD also achieves a substantial ProofBench gain, rising from 25.5 to 34.2, although its AnswerBench score decreases from 68.8 to 67.5.

These results demonstrate the effectiveness of our method for cross-family distillation, while also highlighting the challenge posed by tokenizer mismatch.
SU-01 is based on the Qwen model family and uses a byte-level BPE tokenizer. GLM-4.7-Flash also adopts a BPE-based tokenizer, although with a different vocabulary, whereas Gemma uses a SentencePiece-based tokenizer with substantially different vocabulary and segmentation behavior. Accordingly, GLM benefits consistently across both benchmarks, while the more tokenizer-dissimilar Gemma merely improves on ProofBench. Nevertheless, the clear ProofBench improvements for both models demonstrate that our approach can transfer proof-reasoning capabilities across model families and tokenizer boundaries, although larger tokenizer discrepancies may make such transfer more challenging.

\section{Analysis}
\subsection{Comparison with OPD Baselines}

\begin{figure}[t]
    \centering
    \includegraphics[width=\linewidth]{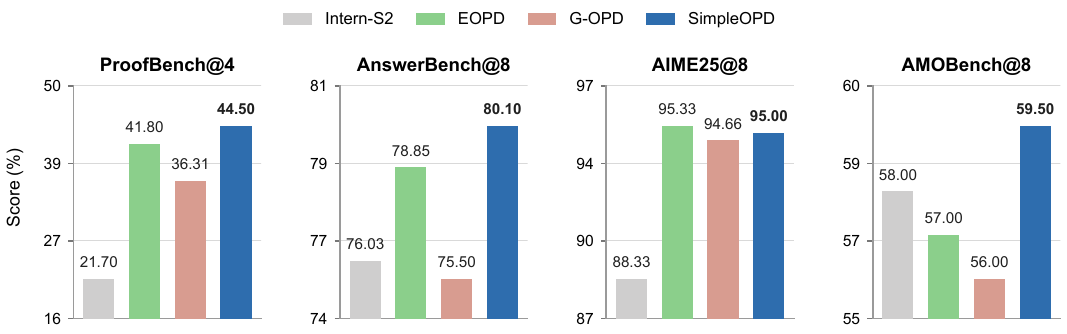}
    \caption{Comparison of the Intern-S2 base model distilled from SU-01 with EOPD, G-OPD, and SimpleOPD across four reasoning benchmarks. All scores are reported as percentages. Each panel uses an independently scaled vertical axis to make small performance differences visible.}
    \label{fig:interns2-opd-baseline-comparison}
\end{figure}

We compare SimpleOPD with the unmodified Intern-S2 model and two recent OPD variants: EOPD, which supplements reverse-KL OPD with forward KL at positions where the teacher has high token-level entropy~\citep{jin2026entropyaware}, and G-OPD, which generalizes OPD through a flexible reference model and reward scaling~\citep{yang2026learningbeyond}. As shown in Figure~\ref{fig:interns2-opd-baseline-comparison}, SimpleOPD improves the Intern-S2 base model by 22.80, 4.07, 6.67, and 1.50 points on ProofBench, AnswerBench, AIME25, and AMOBench, respectively. Among the OPD methods, SimpleOPD obtains 44.50 on ProofBench, 80.10 on AnswerBench, and 59.50 on AMOBench, outperforming both EOPD and G-OPD by a large margin. On AIME25, SimpleOPD reaches 95.00, which is 0.33 points below EOPD and 0.34 points above G-OPD. Overall, SimpleOPD achieves the best result on three of the four benchmarks, with its largest advantage appearing on the proof-oriented ProofBench, while remaining competitive on AIME25.

\subsection{Lexical Overlap}
\begin{figure}
    \centering
    \includegraphics[width=\linewidth]{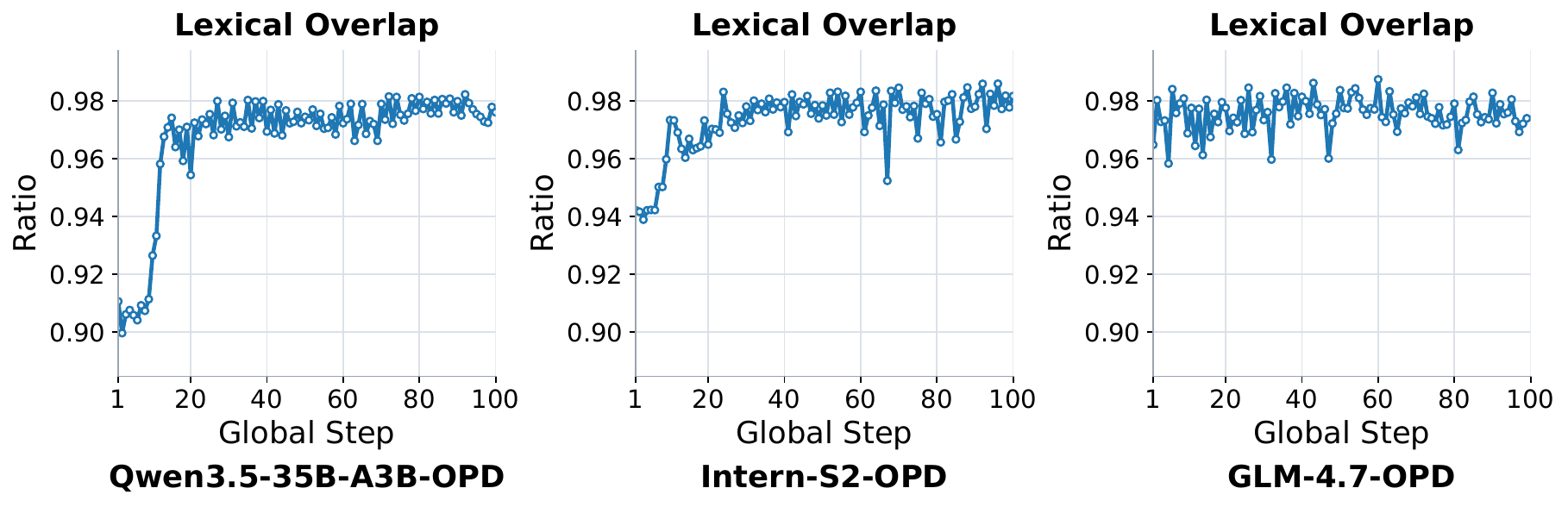}
    \caption{The lexical-overlap curves of Qwen3.5-35B-A3B, Intern-S2-Preview and GLM-4.7-Flash during OPD training from the teacher SU-01.}
    \label{fig:lexical_overlap}
\end{figure}

The lexical-overlap curves provide an estimate of how much token-level supervision can be retained under cross-tokenizer alignment. As shown in Figure \ref{fig:lexical_overlap}, the models already exhibit a high aligned-token ratio at the beginning of training, and the ratio increases further over time. This indicates that, despite tokenizer differences, a large portion of the student-generated text can be matched to teacher tokens through shared surface spans. These results suggest that the partial alignment used in cross-tokenizer OPD preserves a substantial amount of usable training signal in practice, while avoiding the requirement of full tokenizer compatibility.

\subsection{Out-of-domain Generalization}
Although OPD training is conducted only on mathematical reasoning data, the resulting model also shows consistent gains on out-of-domain scientific reasoning benchmarks. The results are shown in Table \ref{tab:science_results}. Compared with Intern-S2-Preview, Intern-S2-OPD achieves a slight improvement on FrontierScience-Olympiad and HLE~\citep{wang2026frontierscience,phan2025humanity}, and a clearer gain on HiPhO~\citep{yu2025hipho}, improving from 38.6 to 41.1. On FrontierScience Research~\citep{wang2026frontierscience}, the score also improves from 1.7 to 5.0. This suggests that the OPD-trained model does not lose its broader scientific reasoning ability despite being trained only on math data, and may further benefit physics-oriented reasoning. In particular, Intern-S2-OPD outperforms SU-01 on HiPhO, indicating that the student model retains its own strengths while gaining additional reasoning capability through SimpleOPD.

\begin{table}[t]
    \centering
    \caption{Evaluation results of Intern-S2-OPD distilled from SU-01 on scientific reasoning benchmarks.}
    \label{tab:science_results}
    \small
    \setlength{\tabcolsep}{3pt}
    \renewcommand{\arraystretch}{1.15}
    \begin{tabular}{lcccc}
        \toprule
        Model
            & \makecell{FrontierScience\\Olympiad}
            & \makecell{FrontierScience\\Research}
            & \makecell{HLE\\(text-only)}
            & HiPhO \\
        \midrule
        SU-01
            & 61.5 & 11.7 & 20.7 & 35.0 \\
        Intern-S2-Preview
            & 60.6 & 1.7 & 19.6 & 38.6 \\
        \rowcolor{opdblue}
        Intern-S2-OPD
            & \textbf{60.9} \deltaup{0.3}
            & \textbf{5.0} \deltaup{3.3}
            & \textbf{20.5} \deltaup{0.9}
            & \textbf{41.1} \deltaup{2.5} \\
        \bottomrule
    \end{tabular}
\end{table}

\subsection{Effect of Training Data}
We further analyze the effect of data composition in distillation. Specifically, we include the verifiable math data used in SU-01 during the OPD process in addition to the proof data.

The results show that both data compositions substantially improve Intern-S2-Preview over the base model, confirming the effectiveness of OPD distillation. Using only proof data achieves the best performance on ProofBench@4, improving Intern-S2 from 21.70 to 44.50 and approaching the teacher model SU-01. This suggests that proof-focused data is more beneficial for transferring natural-language proof-reasoning ability.
Adding verifiable math data brings only a marginal improvement on AnswerBench@8, increasing the score from 80.10 to 81.10, while AIME25@8 remains unchanged at 95.00. However, this mixed-data setting significantly underperforms proof-only data on ProofBench@4, dropping from 44.50 to 38.50. These results indicate that adding verifiable data provides limited additional benefit for general mathematical reasoning, while potentially weakening the transfer of proof-reasoning ability.

\subsection{Effect of OPD Length}
We analyze the effect of the distillation length on the student models (Table \ref{tab:length_ablation}). By comparison, we find that for long-context proof-reasoning tasks, a 6k distillation length is insufficient to fully capture the teacher model’s reasoning patterns. Specifically, increasing the distillation length from 6k to 32k improves the student model consistently across all evaluated benchmarks when the distillation is stable. In Qwen3.5-OPD, ProofBench@4 increases from 40.07 to 42.39, AnswerBench@8 from 77.97 to 80.16, and AIME25@8 from 96.25 to 96.67. The improvement is most pronounced on ProofBench, suggesting that longer distillation sequences are particularly important for transferring complex, multi-step proof reasoning behavior. These results indicate that preserving longer reasoning traces during distillation helps the student model better approximate the teacher’s long-context reasoning capability.

\begin{table}[t]
    \centering
    \caption{Results of different distillation lengths on Intern-S2-Preview and Qwen3.5-35B-A3B distilled from the teacher SU-01.}
    \label{tab:length_ablation}
    \small
    \renewcommand{\arraystretch}{1.1}
    \begin{tabular}{lcccc}
        \toprule
        Model & Length & ProofBench@4 & AnswerBench@8 & AIME25@8 \\
        \midrule
        SU-01
            & -- & 45.00 & 77.50 & 94.60 \\
        Intern-S2-Preview & --
            & 21.70 & 76.03 & 88.33  \\

        Qwen3.5-35B-A3B
            & -- & 26.78 & 73.16 & 94.60 \\ \rowcolor{opdblue}
         & 6k
         & 38.80 \deltaup{17.10}
         & 77.25 \deltaup{1.22}
         & 95.00 \deltaup{6.67}
        
        \\ \rowcolor{opdblue} \multirow{-2}{*}{Intern-S2-OPD} & 32k 
            & \textbf{44.50} \deltaup{22.80}
            & \textbf{80.10} \deltaup{4.07}
            & \textbf{95.00} \deltaup{6.67}
             \\
            
        \rowcolor{opdblue}
            & 6k
            & 40.07 \deltaup{13.29}
            & 77.97 \deltaup{4.81}
            & 96.25 \deltaup{1.65} \\

        \rowcolor{opdblue}
        \multirow{-2}{*}{Qwen3.5-35B-A3B-OPD}
            & 32k
            & \textbf{42.39} \deltaup{15.61}
            & \textbf{80.16} \deltaup{7.00}
            & \textbf{96.67} \deltaup{2.07} \\
        \bottomrule
    \end{tabular}
\end{table}

\subsection{Distillation from DeepSeek-V4-Flash}
Table~\ref{tab:deepseek-intern} reports the results of distilling DeepSeek-V4-Flash \citep{deepseekai2026deepseekv4} into Intern-S2-Preview using a 6K distillation length for computational efficiency, together with termination-token masking and student-reference KL regularization. Despite the shorter distillation length, Intern-S2-DS-OPD consistently outperforms its base model across all three benchmarks. Its ProofBench@4 score increases from 21.70 to 39.71, corresponding to a gain of 18.01 points. AnswerBench@8 improves from 76.03 to 77.94, while AIME25@8 increases from 88.33 to 97.50, yielding gains of 1.91 and 9.17 points, respectively. Moreover, its ProofBench@4 score of 39.71 is 0.91 points higher than the 38.80 achieved using SU-01 as the teacher with the same 6K distillation length in Table~\ref{tab:length_ablation}, suggesting that a stronger teacher can lead to greater improvements in the student. Overall, these results demonstrate that SimpleOPD can effectively transfer reasoning capabilities across distinct model families and from a substantially larger teacher (158B) to a smaller student (30B), even with a relatively short distillation length.


\subsection{Effect of Student-Reference KL Loss}
We further investigate how the student-reference KL coefficient affects GLM distillation. This coefficient controls the extent to which the student is regularized toward its reference policy and therefore influences the balance between preserving the base model's behavior and learning from the teacher. We evaluate coefficients of 0.5, 1.0, and 1.2 to study this trade-off.

As shown in Table~\ref{tab:student_kl}, GLM-4.7-Flash-OPD consistently outperforms the base model under all three settings. Among them, a coefficient of 1.0 achieves the best overall performance, improving ProofBench@4 from 30.75 to 39.71 and AnswerBench@8 from 69.59 to 71.97, with gains of 8.96 and 2.38 points, respectively. A smaller coefficient of 0.5 provides insufficient regularization and yields lower ProofBench and AnswerBench scores. Conversely, increasing the coefficient to 1.2 imposes a stronger constraint on the student policy, which may limit its ability to absorb the teacher's reasoning capabilities. Although coefficients of 0.5 and 1.2 achieve the highest AIME25@8 score of 94.17, their overall performance is less balanced than that obtained with a coefficient of 1.0. These results suggest that the student-reference KL coefficient should be neither too small nor too large, and that a moderate value provides the best balance between retaining the student's original capabilities and acquiring knowledge from the teacher.


\begin{table}[t]
    \centering
    \caption{SimpleOPD results from DeepSeek-V4-Flash to Intern-S2-Preview with masking termination tokens and adding student reference KL loss. The distillation length is 6k for computational efficiency. }
    \label{tab:deepseek-intern}
    \small
    \renewcommand{\arraystretch}{1.1}
    \begin{tabular}{lccc}
        \toprule
        Model & ProofBench@4 & AnswerBench@8 & AIME25@8 \\
        \midrule
        DeepSeek-V4-Flash & 52.18 & 77.00 & 99.17 \\
        Intern-S2-Preview & 21.70 & 76.03 & 88.33  \\
        \rowcolor{opdblue}
Intern-S2-DS-OPD
    & \textbf{39.71} \deltaup{18.01}
    & \textbf{77.94} \deltaup{1.91}
    & \textbf{97.50} \deltaup{9.17} \\
        \bottomrule
    \end{tabular}
\end{table}

\begin{table}[t]
    \centering
    \caption{Ablation study on the student-reference KL coefficient in the setting from SU-01 to GLM-4.7-Flash. Green values indicate improvements over the GLM-4.7-Flash base model.}
    \label{tab:student_kl}
    \small
    \setlength{\tabcolsep}{3pt}
    \renewcommand{\arraystretch}{1.1}
    \begin{tabular}{lcccc}
        \toprule
        Model
            & \makecell{Student KL\\Coefficient}
            & ProofBench@4
            & AnswerBench@8
            & AIME25@8 \\
        \midrule
        SU-01
            & -- & 45.00 & 77.50 & 94.60 \\
        GLM-4.7-Flash
            & -- & 30.75 & 69.59 & 92.08 \\
        \rowcolor{opdblue}
            & 0.5
            & 37.15 \deltaup{6.40}
            & 71.28 \deltaup{1.69}
            & \textbf{94.17} \deltaup{2.09} \\
        \rowcolor{opdblue}
            & 1.0
            & \textbf{39.71} \deltaup{8.96}
            & \textbf{71.97} \deltaup{2.38}
            & 92.91 \deltaup{0.83} \\
        \rowcolor{opdblue}
        \multirow{-3}{*}{GLM-4.7-Flash-OPD}
            & 1.2
            & 37.96 \deltaup{7.21}
            & 71.19 \deltaup{1.60}
            & \textbf{94.17} \deltaup{2.09} \\
        \bottomrule
    \end{tabular}
\end{table}

\section{Related Work}
\label{sec:related-work}
\subsection{Olympiad-Level Long-Context Reasoning}
Early works such as OpenAI o1 and DeepSeek-R1 established long chain-of-thought reasoning as a central test-time scaling paradigm~\citep{openai2024o1systemcard,guo2025deepseekr1}. This approach has enabled LLMs to tackle challenging real-world tasks, including long-horizon agentic tasks and Olympiad-level reasoning.
Recent studies have further advanced reasoning capabilities in mathematics and science. Gemini Deep Think was among the first systems to achieve gold-medal-level performance at the IMO \citep{googledeepmind2025imo}. Nemotron-Math-v2 and MiniMax subsequently reported similarly strong results on mathematical Olympiad benchmarks \citep{du2025nemotronmath,chen2026maxproof}. Beyond mathematics, P1 and P1-VL achieved gold-medal-level performance across several physics Olympiads, while SU-01 introduced a unified recipe for eliciting Olympiad-level proof behavior from a 30B model \citep{chen2025p1,luo2026p1vl,li2026su01}.
However, these capabilities typically rely on costly SFT and RL pipelines, including long-trajectory curation, repeated data-selection rollouts, and RL training. The cost of each stage increases substantially with reasoning length. This motivates our exploration of on-policy distillation for efficiently transferring Olympiad-level proof capabilities across diverse models.
\subsection{On-policy Distillation}
\label{sec:related-opd}
Recently, on-policy distillation (OPD), introduced by MiniLLM and GKD \citep{gu2024minillm,agarwal2024onpolicy}, has emerged as an important component of LLM post-training and has been adopted by flagship models such as GLM-5, DeepSeek-V4, and Qwen3.5 \citep{zeng2026glm5,deepseekai2026deepseekv4,qwenteam2026qwen35}. By optimizing the reverse KL divergence between teacher and student distributions over student-generated rollouts, OPD combines on-policy exploration with dense token-level supervision, offering a stable and generalizable alternative to conventional SFT and RL \citep{gu2024minillm,agarwal2024onpolicy}. Subsequent work has broadened OPD beyond standard autoregressive students: Draft-OPD adapts it to training-based speculative draft models through verification-error replay \citep{lei2026draftopdonpolicydistillationspeculative}, while \citet{shen2026geometryopd} show that OPD follows a characteristic low-dimensional update geometry distinct from SFT and RLVR.

However, its effectiveness can be limited by mismatched reasoning patterns between teacher and student and unreliable supervision over long sequences \citep{agarwal2024onpolicy,lei2026draftopdonpolicydistillationspeculative}. Moreover, most existing studies assume a shared vocabulary, typically distilling models within the same family or combining domain-specific RL teachers \citep{gu2024minillm,agarwal2024onpolicy,deepseekai2026deepseekv4}. OPD between heterogeneous models with different architectures and vocabularies remains underexplored. In this work, we investigate heterogeneous OPD for transferring long-context reasoning capabilities across diverse models.




\section{Conclusion}
\label{sec:conclusion}
In this work, we studied on-policy distillation from a long-context reasoning model, SU-01, to short-context student models. To handle tokenizer differences, we performed OPD in a shared text space and aligned only tokens that occupy identical text spans under the student and teacher tokenizers. We found that naive distillation led to severe training instability: the student’s output length kept increasing, termination tokens were often missing, and many responses were truncated. To address this issue, we introduced two simple but effective stabilization techniques: (1) masking the OPD loss on structural termination tokens, including ${\scriptstyle <}\,\mathrm{/think}{\scriptstyle >}$ and ${\scriptstyle <}\,\mathrm{|im\_end|}\,{\scriptstyle >}$, (2) adding a reference KL loss to constrain the student policy.
These techniques substantially reduced length explosion and truncation while improving reasoning performance. Experiments on both same-family and different-family student models showed consistent gains in mathematical reasoning ability, with Intern-S2-Preview achieving a 21.2-point improvement on ProofBench. These results demonstrate that our approach provides an effective and generalizable way to transfer long-context reasoning capabilities to short-context models across different tokenizers and model families.

\section*{Acknowledgments}

This work was supported by the Shanghai Artificial Intelligence Laboratory. We are grateful to the authors and open-source communities whose work made this project possible.

\bibliography{paper}

@inproceedings{gu2024minillm,
  title={{MiniLLM}: Knowledge Distillation of Large Language Models},
  author={Gu, Yuxian and Dong, Li and Wei, Furu and Huang, Minlie},
  booktitle={The Twelfth International Conference on Learning Representations},
  year={2024},
  url={https://openreview.net/forum?id=5h0qf7IBZZ}
}

@misc{gemmateam2026gemma4,
      title={Gemma 4 Technical Report}, 
      author={{Gemma Team}},
      year={2026},
      eprint={2607.02770},
      archivePrefix={arXiv},
      primaryClass={cs.CL},
      url={https://arxiv.org/abs/2607.02770}, 
}

@inproceedings{agarwal2024onpolicy,
  title={On-Policy Distillation of Language Models: Learning from Self-Generated Mistakes},
  author={Agarwal, Rishabh and Vieillard, Nino and Zhou, Yongchao and Stanczyk, Piotr and Ramos Garea, Sabela and Geist, Matthieu and Bachem, Olivier},
  booktitle={The Twelfth International Conference on Learning Representations},
  year={2024},
  url={https://openreview.net/forum?id=3zKtaqxLhW}
}

@article{jin2026entropyaware,
  title={Entropy-Aware On-Policy Distillation of Language Models},
  author={Jin, Woogyeol and Min, Taywon and Yang, Yongjin and Wei, Dennis and Zhou, Yi and Kadhe, Swanand Ravindra and Baracaldo, Nathalie and Lee, Kimin},
  journal={arXiv preprint arXiv:2603.07079},
  year={2026},
  url={https://arxiv.org/abs/2603.07079}
}

@article{yang2026learningbeyond,
  title={Learning beyond Teacher: Generalized On-Policy Distillation with Reward Extrapolation},
  author={Yang, Wenkai and Liu, Weijie and Xie, Ruobing and Yang, Kai and Yang, Saiyong and Lin, Yankai},
  journal={arXiv preprint arXiv:2602.12125},
  year={2026},
  url={https://arxiv.org/abs/2602.12125}
}

@article{shen2026geometryopd,
  title={On the Geometry of On-Policy Distillation},
  author={Shen, Zhennan and Li, Yanshu and Yin, Qingyu and Leong, Chak Tou and Wang, Zhilin and Chen, Yanxu and Han, Rongduo and Lee, Sunbowen and Fung, Yi R.},
  journal={arXiv preprint arXiv:2606.07082},
  year={2026},
  url={https://arxiv.org/abs/2606.07082}
}

@misc{lei2026draftopdonpolicydistillationspeculative,
  title={Draft-OPD: On-Policy Distillation for Speculative Draft Models},
  author={Haodi Lei and Yafu Li and Haoran Zhang and Shunkai Zhang and Qianjia Cheng and others},
  year={2026},
  eprint={2605.29343},
  archivePrefix={arXiv},
  primaryClass={cs.CL},
  url={https://arxiv.org/abs/2605.29343},
}

@misc{openai2024o1systemcard,
  title={{OpenAI o1} System Card},
  author={{OpenAI}},
  year={2024},
  url={https://openai.com/index/openai-o1-system-card/}
}

@article{guo2025deepseekr1,
  title={{DeepSeek-R1}: Incentivizing Reasoning Capability in {LLM}s via Reinforcement Learning},
  author={{DeepSeek-AI} and Guo, Daya and Yang, Dejian and Zhang, Haowei and Song, Junxiao and Wang, Peiyi and Zhu, Qihao and Xu, Runxin and Zhang, Ruoyu and Ma, Shirong and others},
  journal={Nature},
  volume={645},
  pages={633--638},
  year={2025},
  url={https://arxiv.org/abs/2501.12948}
}

@article{chen2025p1,
  title={{P1}: Mastering Physics Olympiads with Reinforcement Learning},
  author={Chen, Jiacheng and Cheng, Qianjia and Yu, Fangchen and Wan, Haiyuan and Zhang, Yuchen and Zheng, Shenghe and Yao, Junchi and Zhang, Qingyang and He, Haonan and Luo, Yun and others},
  journal={arXiv preprint arXiv:2511.13612},
  year={2025},
  url={https://arxiv.org/abs/2511.13612}
}

@article{li2026su01,
  title={Achieving Gold-Medal-Level Olympiad Reasoning via Simple and Unified Scaling},
  author={Li, Yafu and Zhan, Runzhe and Zhang, Haoran and Zhang, Shunkai and Li, Yizhuo and Wang, Zhilin and Chen, Jiacheng and Wang, Futing and Hu, Xuyang and Fan, Yuchen and others},
  journal={arXiv preprint arXiv:2605.13301},
  year={2026},
  url={https://arxiv.org/abs/2605.13301}
}

@inproceedings{zhang-etal-2024-dual,
  title={Dual-Space Knowledge Distillation for Large Language Models},
  author={Zhang, Songming and Zhang, Xue and Sun, Zengkui and Chen, Yufeng and Xu, Jinan},
  booktitle={Proceedings of the 2024 Conference on Empirical Methods in Natural Language Processing},
  pages={18164--18181},
  year={2024},
  address={Miami, Florida, USA},
  publisher={Association for Computational Linguistics},
  doi={10.18653/v1/2024.emnlp-main.1010},
  url={https://aclanthology.org/2024.emnlp-main.1010/}
}

@inproceedings{chen-etal-2025-enhancing-cross,
  title={Enhancing Cross-Tokenizer Knowledge Distillation with Contextual Dynamical Mapping},
  author={Chen, Yijie and Liu, Yijin and Meng, Fandong and Chen, Yufeng and Xu, Jinan and Zhou, Jie},
  booktitle={Findings of the Association for Computational Linguistics: ACL 2025},
  pages={8005--8018},
  year={2025},
  address={Vienna, Austria},
  publisher={Association for Computational Linguistics},
  doi={10.18653/v1/2025.findings-acl.419},
  url={https://aclanthology.org/2025.findings-acl.419/}
}

@inproceedings{singh-etal-2026-cross,
  title={Cross-Tokenizer {LLM} Distillation through a Byte-Level Interface},
  author={Singh, Avyav Kumar and Wu, Yen-Chen and Cioba, Alexandru and Bernacchia, Alberto and Buffelli, Davide},
  booktitle={Proceedings of the Second Workshop on Customizable NLP: Progress and Challenges in Customizing NLP for a Domain, Application, Group, or Individual},
  pages={84--96},
  year={2026},
  address={San Diego, California, USA},
  publisher={Association for Computational Linguistics},
  doi={10.18653/v1/2026.customnlp4u-1.9},
  url={https://aclanthology.org/2026.customnlp4u-1.9/}
}

@misc{slime_github,
  author={Zhu, Zilin and Xie, Chengxing and Lv, Xin and {slime Contributors}},
  title={slime: An {LLM} Post-Training Framework for {RL} Scaling},
  year={2025},
  howpublished={\url{https://github.com/THUDM/slime}},
  note={GitHub repository}
}

@inproceedings{zheng2024sglang,
  title={{SGLang}: Efficient Execution of Structured Language Model Programs},
  author={Zheng, Lianmin and Yin, Liangsheng and Xie, Zhiqiang and Sun, Chuyue and Huang, Jeff and Yu, Cody Hao and Cao, Shiyi and Kozyrakis, Christos and Stoica, Ion and Gonzalez, Joseph E. and Barrett, Clark and Sheng, Ying},
  booktitle={Advances in Neural Information Processing Systems},
  year={2024},
  doi={10.48550/arXiv.2312.07104},
  url={https://arxiv.org/abs/2312.07104}
}

@misc{googledeepmind2025imo,
  title={Advanced Version of {Gemini} with {Deep Think} Officially Achieves Gold-Medal Standard at the International Mathematical Olympiad},
  author={Luong, Thang and Lockhart, Edward},
  year={2025},
  month={jul},
  howpublished={Google DeepMind Blog},
  url={https://deepmind.google/blog/advanced-version-of-gemini-with-deep-think-officially-achieves-gold-medal-standard-at-the-international-mathematical-olympiad/}
}

@article{du2025nemotronmath,
  title={{Nemotron-Math}: Efficient Long-Context Distillation of Mathematical Reasoning from Multi-Mode Supervision},
  author={Du, Wei and Toshniwal, Shubham and Kisacanin, Branislav and Mahdavi, Sadegh and Moshkov, Ivan and Armstrong, George and Ge, Stephen and Minasyan, Edgar and Chen, Feng and Gitman, Igor},
  journal={arXiv preprint arXiv:2512.15489},
  year={2025},
  url={https://arxiv.org/abs/2512.15489}
}

@article{chen2026maxproof,
  title={{MaxProof}: Scaling Mathematical Proof with Generative-Verifier {RL} and Population-Level Test-Time Scaling},
  author={Chen, Jiacheng and Zhang, Xinyu and Zhang, Shunkai and Wang, Yanmohan and Li, Lin and Qin, Tiancheng and Wang, Qin and Zhu, Zhengmao and Li, Tianle and Li, Jingyang and others},
  journal={arXiv preprint arXiv:2606.13473},
  year={2026},
  url={https://arxiv.org/abs/2606.13473}
}

@article{luo2026p1vl,
  title={{P1-VL}: Bridging Visual Perception and Scientific Reasoning in Physics Olympiads},
  author={Luo, Yun and Wang, Futing and Cheng, Qianjia and Yu, Fangchen and Lei, Haodi and Yan, Jianhao and Li, Chenxi and Chen, Jiacheng and Zhao, Yufeng and Wan, Haiyuan and others},
  journal={arXiv preprint arXiv:2602.09443},
  year={2026},
  url={https://arxiv.org/abs/2602.09443}
}

@article{zeng2026glm5,
  title={{GLM-5}: From Vibe Coding to Agentic Engineering},
  author={{GLM-5 Team}},
  journal={arXiv preprint arXiv:2602.15763},
  year={2026},
  url={https://arxiv.org/abs/2602.15763}
}

@misc{deepseekai2026deepseekv4,
  title={{DeepSeek-V4}: Towards Highly Efficient Million-Token Context Intelligence},
  author={{DeepSeek-AI}},
  year={2026},
  note={Technical report},
  url={https://huggingface.co/deepseek-ai/DeepSeek-V4-Pro/blob/main/DeepSeek_V4.pdf}
}

@misc{qwenteam2026qwen35,
  title={{Qwen3.5}},
  author={{Qwen Team}},
  year={2026},
  howpublished={Model release},
  url={https://qwen.ai/blog?id=qwen3.5}
}

@misc{sun2026simct,
  title={{SimCT}: Recovering Lost Supervision for Cross-Tokenizer On-Policy Distillation},
  author={Jie Sun and Mao Zheng and Mingyang Song and Qiyong Zhong and Yilin Cheng and Bichuan Feng and Pengfei Liu and Junfeng Fang and Xiang Wang},
  year={2026},
  eprint={2605.07711},
  archivePrefix={arXiv},
  primaryClass={cs.CL},
  url={https://arxiv.org/abs/2605.07711}
}

@article{yang2025qwen3,
  title={{Qwen3} Technical Report},
  author={Yang, An and Li, Anfeng and Yang, Baosong and Zhang, Beichen and Hui, Binyuan and Zheng, Bo and Yu, Bowen and Gao, Chang and Huang, Chengen and Lv, Chenxu and others},
  journal={arXiv preprint arXiv:2505.09388},
  year={2025},
  url={https://arxiv.org/abs/2505.09388}
}

@misc{internlm2026interns2preview,
  title={{Intern-S2-Preview}},
  author={{InternLM Team}},
  year={2026},
  howpublished={Hugging Face model card},
  url={https://huggingface.co/internlm/Intern-S2-Preview}
}

@misc{zai2026glm47flash,
  title={{GLM-4.7-Flash}},
  author={{Z.ai}},
  year={2026},
  howpublished={Hugging Face model card},
  url={https://huggingface.co/zai-org/GLM-4.7-Flash}
}

@article{li2026rethinkingopd,
  title={Rethinking On-Policy Distillation of Large Language Models: Phenomenology, Mechanism, and Recipe},
  author={Li, Yaxuan and Zuo, Yuxin and He, Bingxiang and Zhang, Jinqian and Xiao, Chaojun and Qian, Cheng and Yu, Tianyu and Gao, Huan-ang and Yang, Wenkai and Liu, Zhiyuan and Ding, Ning},
  journal={arXiv preprint arXiv:2604.13016},
  year={2026},
  url={https://arxiv.org/abs/2604.13016}
}

@inproceedings{luong-etal-2025-towards,
  title={Towards Robust Mathematical Reasoning},
  author={Luong, Thang and Hwang, Dawsen and Nguyen, Hoang H. and Ghiasi, Golnaz and Chervonyi, Yuri and Seo, Insuk and Kim, Junsu and Bingham, Garrett and Lee, Jonathan and Mishra, Swaroop and Zhai, Alex and Hu, Huiyi and Michalewski, Henryk and Kim, Jimin and Ahn, Jeonghyun and Bae, Junhwi and Song, Xingyou and Trinh, Trieu Hoang and Le, Quoc V. and Jung, Junehyuk},
  booktitle={Proceedings of the 2025 Conference on Empirical Methods in Natural Language Processing},
  pages={35418--35442},
  year={2025},
  address={Suzhou, China},
  publisher={Association for Computational Linguistics},
  doi={10.18653/v1/2025.emnlp-main.1794},
  url={https://aclanthology.org/2025.emnlp-main.1794/}
}

@article{phan2025humanity,
  title={Humanity's Last Exam},
  author={Phan, Long and Gatti, Alice and Han, Ziwen and Li, Nathaniel and Hu, Josephina and Zhang, Hugh and Zhang, Chen Bo Calvin and Shaaban, Mohamed and Ling, John and Shi, Sean and others},
  journal={arXiv preprint arXiv:2501.14249},
  year={2025},
  url={https://arxiv.org/abs/2501.14249}
}

@article{yu2025hipho,
  title={{HiPhO}: How Far Are ({M}){LLM}s from Humans in the Latest High School Physics Olympiad Benchmark?},
  author={Yu, Fangchen and Wan, Haiyuan and Cheng, Qianjia and Zhang, Yuchen and Chen, Jiacheng and Han, Fujun and Wu, Yulun and Yao, Junchi and Hu, Ruilizhen and Ding, Ning and Cheng, Yu and Chen, Tao and Bai, Lei and Zhou, Dongzhan and Luo, Yun and Cui, Ganqu and Ye, Peng},
  journal={arXiv preprint arXiv:2509.07894},
  year={2025},
  url={https://arxiv.org/abs/2509.07894}
}

@article{an2025amobench,
  title={{AMO-Bench}: Large Language Models Still Struggle in High School Math Competitions},
  author={An, Shengnan and Cai, Xunliang and Cao, Xuezhi and Li, Xiaoyu and Lin, Yehao and Liu, Junlin and Lv, Xinxuan and Ma, Dan and Wang, Xuanlin and Wang, Ziwen and Zhou, Shuang},
  journal={arXiv preprint arXiv:2510.26768},
  year={2025},
  url={https://arxiv.org/abs/2510.26768}
}

@misc{aime2025,
  title={American Invitational Mathematics Examination (AIME) 2025},
  author={{Mathematical Association of America}},
  year={2025},
  url={https://maa.org/math-competitions/american-invitational-mathematics-examination-aime}
}

@article{wang2026frontierscience,
  title={{FrontierScience}: Evaluating {AI}'s Ability to Perform Expert-Level Scientific Tasks},
  author={Wang, Miles and Lin, Robi and Hu, Kat and Jiao, Joy and Chowdhury, Neil and Chang, Ethan and Patwardhan, Tejal},
  journal={arXiv preprint arXiv:2601.21165},
  year={2026},
  url={https://arxiv.org/abs/2601.21165}
}

@article{lu2025onpolicydistillation,
  author = {Kevin Lu and Thinking Machines Lab},
  title = {On-Policy Distillation},
  journal = {Thinking Machines Lab: Connectionism},
  year = {2025},
  note = {https://thinkingmachines.ai/blog/on-policy-distillation},
  doi = {10.64434/tml.20251026},
}

@misc{ma2026mopdmultiteacheronpolicydistillation,
      title={MOPD: Multi-Teacher On-Policy Distillation for Capability Integration in LLM Post-Training}, 
      author={Wenhan Ma and Jianyu Wei and Liang Zhao and Hailin Zhang and Bangjun Xiao and Lei Li and Qibin Yang and Bofei Gao and Yudong Wang and Rang Li and Jinhao Dong and Zhifang Sui and Fuli Luo},
      year={2026},
      eprint={2606.30406},
      archivePrefix={arXiv},
      primaryClass={cs.CL},
      url={https://arxiv.org/abs/2606.30406}, 
}

@misc{hübotter2026reinforcementlearningselfdistillation,
      title={Reinforcement Learning via Self-Distillation}, 
      author={Jonas Hübotter and Frederike Lübeck and Lejs Behric and Anton Baumann and Marco Bagatella and Daniel Marta and Ido Hakimi and Idan Shenfeld and Thomas Kleine Buening and Carlos Guestrin and Andreas Krause},
      year={2026},
      eprint={2601.20802},
      archivePrefix={arXiv},
      primaryClass={cs.LG},
      url={https://arxiv.org/abs/2601.20802}, 
}

@misc{openai2025gptoss120bgptoss20bmodel,
      title={gpt-oss-120b \& gpt-oss-20b Model Card}, 
      author={OpenAI and : and Sandhini Agarwal and Lama Ahmad and Jason Ai and Sam Altman and Andy Applebaum and Edwin Arbus and Rahul K. Arora and Yu Bai and Bowen Baker and Haiming Bao and Boaz Barak and Ally Bennett and Tyler Bertao and Nivedita Brett and Eugene Brevdo and Greg Brockman and Sebastien Bubeck and Che Chang and Kai Chen and Mark Chen and Enoch Cheung and Aidan Clark and Dan Cook and Marat Dukhan and Casey Dvorak and Kevin Fives and Vlad Fomenko and Timur Garipov and Kristian Georgiev and Mia Glaese and Tarun Gogineni and Adam Goucher and Lukas Gross and Katia Gil Guzman and John Hallman and Jackie Hehir and Johannes Heidecke and Alec Helyar and Haitang Hu and Romain Huet and Jacob Huh and Saachi Jain and Zach Johnson and Chris Koch and Irina Kofman and Dominik Kundel and Jason Kwon and Volodymyr Kyrylov and Elaine Ya Le and Guillaume Leclerc and James Park Lennon and Scott Lessans and Mario Lezcano-Casado and Yuanzhi Li and Zhuohan Li and Ji Lin and Jordan Liss and Lily and Liu and Jiancheng Liu and Kevin Lu and Chris Lu and Zoran Martinovic and Lindsay McCallum and Josh McGrath and Scott McKinney and Aidan McLaughlin and Song Mei and Steve Mostovoy and Tong Mu and Gideon Myles and Alexander Neitz and Alex Nichol and Jakub Pachocki and Alex Paino and Dana Palmie and Ashley Pantuliano and Giambattista Parascandolo and Jongsoo Park and Leher Pathak and Carolina Paz and Ludovic Peran and Dmitry Pimenov and Michelle Pokrass and Elizabeth Proehl and Huida Qiu and Gaby Raila and Filippo Raso and Hongyu Ren and Kimmy Richardson and David Robinson and Bob Rotsted and Hadi Salman and Suvansh Sanjeev and Max Schwarzer and D. Sculley and Harshit Sikchi and Kendal Simon and Karan Singhal and Yang Song and Dane Stuckey and Zhiqing Sun and Philippe Tillet and Sam Toizer and Foivos Tsimpourlas and Nikhil Vyas and Eric Wallace and Xin Wang and Miles Wang and Olivia Watkins and Kevin Weil and Amy Wendling and Kevin Whinnery and Cedric Whitney and Hannah Wong and Lin Yang and Yu Yang and Michihiro Yasunaga and Kristen Ying and Wojciech Zaremba and Wenting Zhan and Cyril Zhang and Brian Zhang and Eddie Zhang and Shengjia Zhao},
      year={2025},
      eprint={2508.10925},
      archivePrefix={arXiv},
      primaryClass={cs.CL},
      url={https://arxiv.org/abs/2508.10925}, 
}
\bibliographystyle{iclr2026_conference}

\clearpage
\appendix
\section{Cross-Tokenizer Scoring and Alignment}
\label{app:cross-tokenizer-alignment}
The complete procedure of cross-tokenizer scoring and alignment is shown in Algorithm 1.
\begin{algorithm}[hp]
\small
\DontPrintSemicolon
\SetAlgoLongEnd
\SetAlgoCaptionSeparator{\ }
\SetKwInOut{Inputs}{Inputs}
\SetKwInOut{Outputs}{Outputs}
\caption{Cross-tokenizer teacher scoring and greedy token-piece alignment}
\label{alg:cross-tokenizer-alignment}
\Inputs{message sequence $\mathbf{x}$; student response tokens $y_{1:n}$; student log-probabilities $\ell_{1:n}^{\theta}$.}
\Outputs{aligned teacher log-probabilities $\widetilde{\ell}_{1:n}^{\phi}$; matched set $\mathcal{M}$.}

\textcolor{deltagreen}{$\triangleright$\ \textbf{Token-out, text-in teacher scoring}}
\;
\textcolor{blue}{$s \leftarrow \mathcal{D}_{\theta}(y_{1:n}),\quad c_{\phi} \leftarrow \mathcal{C}_{\phi}(\mathbf{x}),\quad u_{\phi} \leftarrow c_{\phi} \oplus s$}
\;
\textcolor{blue}{$z=(z_1,\ldots,z_m) \leftarrow \mathcal{E}_{\phi}(s \mid c_{\phi})$}
\;
\textcolor{blue}{$\ell_i^{\phi} \leftarrow \log \pi_{\phi}(z_i \mid c_{\phi},z_{<i}),\quad \forall i \in \{1,\ldots,m\}$}
\;

\textcolor{deltagreen}{$\triangleright$\ \textbf{Initialize all positions with student log-probabilities}}
\;
\textcolor{orange}{$\widetilde{\ell}_{1:n}^{\phi} \leftarrow \ell_{1:n}^{\theta},\quad \mathcal{M} \leftarrow \emptyset$}
\;
$i \leftarrow 1,\quad j \leftarrow 1,\quad H_{\phi} \leftarrow \varepsilon,\quad H_{\theta} \leftarrow \varepsilon$
\;

\textcolor{deltagreen}{$\triangleright$\ \textbf{$\tau_{\phi}$ and $\tau_{\theta}$ return tokenizer-native token pieces}}
\;
\While{$i \leq m \land j \leq n$}{
    $p_{\phi} \leftarrow \tau_{\phi}(z_i),\quad p_{\theta} \leftarrow \tau_{\theta}(y_j)$
    \;
    \uIf{\textcolor{red}{$H_{\phi}=H_{\theta} \land p_{\phi}=p_{\theta}$}}{
        \textcolor{purple}{$\widetilde{\ell}_{j}^{\phi} \leftarrow \ell_i^{\phi},\quad \mathcal{M} \leftarrow \mathcal{M} \cup \{(i,j)\}$}
        \;
        $H_{\phi} \leftarrow H_{\phi} \oplus p_{\phi},\quad H_{\theta} \leftarrow H_{\theta} \oplus p_{\theta},\quad i \leftarrow i+1,\quad j \leftarrow j+1$
        \;
    }
    \uElseIf{$\lvert H_{\phi}\rvert > \lvert H_{\theta}\rvert$}{
        $H_{\theta} \leftarrow H_{\theta} \oplus p_{\theta},\quad j \leftarrow j+1$
        \;
    }
    \uElseIf{$\lvert H_{\phi}\rvert < \lvert H_{\theta}\rvert$}{
        $H_{\phi} \leftarrow H_{\phi} \oplus p_{\phi},\quad i \leftarrow i+1$
        \;
    }
    \Else{
        $H_{\phi} \leftarrow H_{\phi} \oplus p_{\phi},\quad H_{\theta} \leftarrow H_{\theta} \oplus p_{\theta},\quad i \leftarrow i+1,\quad j \leftarrow j+1$
        \;
    }
}
\textbf{return} $\widetilde{\ell}_{1:n}^{\phi},\mathcal{M}$
\;
\end{algorithm}

\begin{figure}[hp]
    \centering
    \includegraphics[width=\linewidth]{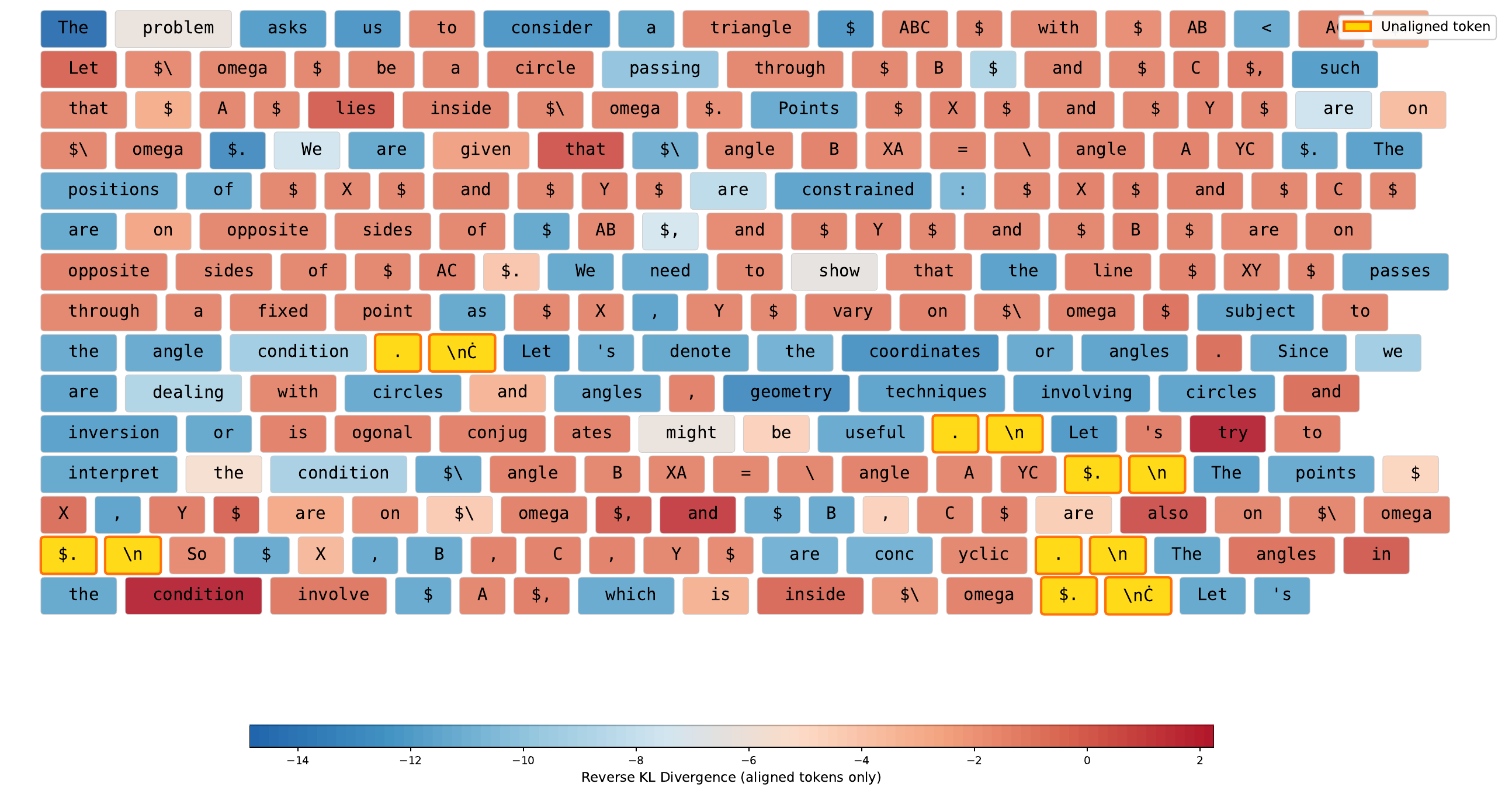}
    \caption{Per-token reverse KL divergence in cross-vocabulary distillation. Colored boxes show aligned tokens with KL divergence values, while gold boxes indicate tokens that could not be aligned between teacher and student models.}
    \label{fig:token_kl}
\end{figure}

As shown in Figure~\ref{fig:token_kl}, which plots an example of cross-tokenizer scoring in the distillation from SU-01 to Intern-S2-Preview, the visualization confirms that our alignment process allows most semantically informative tokens to receive supervision from the teacher.

\clearpage
\section{Model Solutions for ProofBench}
\label{app:proofbench-solutions}
This section presents two ProofBench examples to provide a qualitative view of the mathematical proof-generation behavior learned through SimpleOPD. Both ProofBench-Basic-001 and ProofBench-Advanced-028 are outputs generated by Intern-S2-OPD, the Intern-S2-Preview student distilled from SU-01; each case includes the original problem statement, a concise solution overview, and the complete model-generated proof.
\subsection{ProofBench-Basic-001}
\label{case:proofbench-basic-001}

\newcommand{\proofbenchstep}[1]{%
  \par\medskip
  \noindent\textbf{#1}\par\nobreak\smallskip
}

\begin{casebox}{Question}
Determine all functions $f\colon \mathbb{Z}\to\mathbb{Z}$ such that, for all $x,y\in\mathbb{Z}$,
\[
f(2x)+2f(y)=f(f(x+y)).
\]
\end{casebox}

\begin{proofsolutionbox}{Solution Overview}
\small
The solution first evaluates the equation at carefully chosen arguments to derive a doubling identity and a translated Cauchy relation. Subtracting the constant $f(0)$ then reduces the problem to an additive function on $\mathbb{Z}$, which must be linear. Comparing coefficients and directly verifying the resulting candidates completes the classification.
\end{proofsolutionbox}

\begingroup
\setcounter{equation}{0}
\renewcommand{\theequation}{\thesection.\arabic{equation}}
\renewcommand{\theHequation}{proofbenchbasic001.\arabic{equation}}
\tcbset{title after break={Intern-S2-OPD Solution (continued)}}
\begin{proofsolutionbox}{Intern-S2-OPD Solution}
\small
\noindent\textbf{Solution.}\par\smallskip
For all $x,y\in\mathbb{Z}$, let $P(x,y)$ denote the assertion
\begin{equation}
f(2x)+2f(y)=f(f(x+y))
\label{eq:proofbench-fe}
\end{equation}

\proofbenchstep{Step 1. Two useful identities.}
For every $t\in\mathbb{Z}$, $P(0,t)$ and $P(t,0)$ give, respectively,
\begin{align}
f(f(t)) &= f(0)+2f(t), \label{eq:proofbench-at-zero}\\
f(f(t)) &= f(2t)+2f(0). \label{eq:proofbench-zero}
\end{align}
Equating these expressions gives, for every $t\in\mathbb{Z}$,
\begin{equation}
f(2t)=2f(t)-f(0).
\label{eq:proofbench-doubling}
\end{equation}

\proofbenchstep{Step 2. Deriving a Cauchy-type relation.}
Substituting (\ref{eq:proofbench-doubling}) into (\ref{eq:proofbench-fe}) yields
\begin{equation}
f(f(x+y))=2f(x)+2f(y)-f(0).
\label{eq:proofbench-combined}
\end{equation}
On the other hand, applying (\ref{eq:proofbench-at-zero}) with $t$ replaced by $x+y$ gives
\begin{equation}
f(f(x+y))=f(0)+2f(x+y).
\label{eq:proofbench-composed}
\end{equation}
Equating (\ref{eq:proofbench-combined}) and (\ref{eq:proofbench-composed}), we obtain, for all $x,y\in\mathbb{Z}$,
\[
f(0)+2f(x+y)=2f(x)+2f(y)-f(0),
\]
and hence
\begin{equation}
f(x+y)=f(x)+f(y)-f(0)
\label{eq:proofbench-cauchy}
\end{equation}

\proofbenchstep{Step 3. Additivity after subtracting the constant.}
Define $g(x)=f(x)-f(0)$. Then $g(0)=0$, and, for all $x,y\in\mathbb{Z}$, (\ref{eq:proofbench-cauchy}) becomes
\[
g(x+y)=g(x)+g(y).
\]
Thus $g$ is additive on $\mathbb{Z}$. Since $\mathbb{Z}$ is cyclic, any additive map $\mathbb{Z}\to\mathbb{Z}$ is determined by its value at $1$. In particular, for every $n\in\mathbb{Z}$,
\[
g(n)=n g(1).
\]
Let $a=g(1)\in\mathbb{Z}$ and $b=f(0)$. Consequently, for every $x\in\mathbb{Z}$,
\[
f(x)=ax+b.
\]

\proofbenchstep{Step 4. Determining the parameters $a$ and $b$.}
Substituting $f(x)=ax+b$ into (\ref{eq:proofbench-fe}), the left-hand side is
\begin{align*}
f(2x)+2f(y)
&=a(2x)+b+2(ay+b)\\
&=2a(x+y)+3b,
\end{align*}
while the right-hand side is
\begin{align*}
f(f(x+y))
&=f(a(x+y)+b)\\
&=a^2(x+y)+ab+b.
\end{align*}
Since $t=x+y$ ranges over $\mathbb{Z}$, for every $t\in\mathbb{Z}$ we have
\[
2at+3b=a^2t+ab+b.
\]
Therefore,
\begin{equation}
\begin{aligned}
2a-a^2&=0, & 2b-ab&=0,\\
a(2-a)&=0, & b(2-a)&=0.
\end{aligned}
\label{eq:proofbench-parameters}
\end{equation}
It follows that $a=0$ or $a=2$.
If $a=0$, then $2b=0$, so $b=0$ and $f\equiv 0$.
If $a=2$, the second condition is automatically satisfied for every $b\in\mathbb{Z}$, giving $f(x)=2x+b$.

\proofbenchstep{Step 5. Verification.}
For $f\equiv 0$, both sides of (\ref{eq:proofbench-fe}) are zero. For $f(x)=2x+b$,
\begin{align*}
f(2x)+2f(y)
&=(4x+b)+2(2y+b)\\
&=4x+4y+3b,
\end{align*}
and
\begin{align*}
f(f(x+y))
&=f(2(x+y)+b)\\
&=4x+4y+3b.
\end{align*}
Hence both families satisfy the equation, and the complete set of solutions is
\[
\boxed{
\begin{aligned}
&f\equiv 0, \quad\text{or}\\
&f(x)=2x+c, \qquad c\in\mathbb{Z}.
\end{aligned}}
\]
\end{proofsolutionbox}
\endgroup

\subsection{ProofBench-Advanced-028}
\label{case:proofbench-advanced-028}

\begin{casebox}{Question}
Let $H$ be the orthocenter of acute triangle $ABC$, let $F$ be the foot of the altitude from $C$ to $AB$, and let $P$ be the reflection of $H$ across $BC$. Suppose that the circumcircle of triangle $AFP$ intersects line $BC$ at two distinct points $X$ and $Y$. Prove that $C$ is the midpoint of $XY$.
\end{casebox}

\begin{proofsolutionbox}{Solution Overview}
\small
The solution places $BC$ on the horizontal axis and computes the coordinates of $F$, $H$, and $P$. It then shows that the circumcenter of $\triangle AFP$ lies on the vertical line through $C$. Since this line is perpendicular to the chord $XY\subset BC$, the standard chord-bisection property of a circle implies that $C$ is the midpoint of $XY$.
\end{proofsolutionbox}

\begingroup
\setcounter{equation}{0}
\renewcommand{\theequation}{\thesubsection.\arabic{equation}}
\renewcommand{\theHequation}{proofbenchadvanced028.\arabic{equation}}
\tcbset{title after break={Intern-S2-OPD Solution (continued)}}
\begin{proofsolutionbox}{Intern-S2-OPD Solution}
\small
\noindent\textbf{Solution.}\par\smallskip

\proofbenchstep{Step 1. Coordinate setup.}
We place the triangle in a convenient coordinate system. Since the statement involves only midpoints, collinearity, and perpendicularity, we may apply a similarity transformation without loss of generality. Thus, let
\begin{equation}
B=(0,0),\qquad C=(1,0),\qquad A=(p,q),
\label{eq:proofbench-advanced-coordinates}
\end{equation}
where $q>0$. Since the triangle is acute, $0<p<1$, although these inequalities are not essential for the algebra below. Define
\[
D=p^2+q^2.
\]

\proofbenchstep{Step 2. Coordinates of the auxiliary points.}
The line $AB$ passes through the origin and has direction $(p,q)$. Therefore, the projection of $C=(1,0)$ onto $AB$ is
\begin{equation}
F=\frac{C\mathbin{\cdot}A}{A\mathbin{\cdot}A}A
 =\frac{p}{p^2+q^2}(p,q)
 =\left(\frac{p^2}{D},\frac{pq}{D}\right).
\label{eq:proofbench-advanced-f}
\end{equation}

Because $BC$ is horizontal, the altitude from $A$ is the vertical line $x=p$. The altitude from $B$ is perpendicular to $AC$. Since $AC=(1-p,-q)$, a perpendicular direction is $(q,1-p)$, and hence this altitude has equation
\[
y=\frac{1-p}{q}x.
\]
Intersecting it with $x=p$ gives
\begin{equation}
H=\left(p,\frac{p(1-p)}{q}\right).
\label{eq:proofbench-advanced-h}
\end{equation}
Reflecting $H$ across $BC$, which is the $x$-axis, yields
\begin{equation}
P=\left(p,-\frac{p(1-p)}{q}\right).
\label{eq:proofbench-advanced-p}
\end{equation}

\proofbenchstep{Step 3. A candidate for the circumcenter of $\triangle AFP$.}
The points $A$ and $P$ have the same $x$-coordinate, so $AP$ is vertical. Its perpendicular bisector is the horizontal line through the midpoint
\begin{align*}
M_{AP}
&=\left(p,\frac{q-\frac{p(1-p)}{q}}{2}\right)\\
&=\left(p,\frac{q^2-p(1-p)}{2q}\right).
\end{align*}
Since
\[
q^2-p(1-p)=p^2+q^2-p=D-p,
\]
define
\begin{equation}
y_0=\frac{D-p}{2q}.
\label{eq:proofbench-advanced-y0}
\end{equation}
The perpendicular bisector of $AP$ is therefore $y=y_0$. We claim that
\begin{equation}
O=(1,y_0)
\label{eq:proofbench-advanced-o}
\end{equation}
is the circumcenter of $\triangle AFP$. By construction, $O$ lies on the perpendicular bisector of $AP$. It remains to show that $O$ is equidistant from $A$ and $F$.

\proofbenchstep{Step 4. Verifying that $OA=OF$.}
From equations~(\ref{eq:proofbench-advanced-coordinates}), (\ref{eq:proofbench-advanced-f}), and (\ref{eq:proofbench-advanced-o}),
\begin{align}
OA^2&=(1-p)^2+(y_0-q)^2,\notag\\[-2pt]
OF^2&=\left(1-\frac{p^2}{D}\right)^2
     +\left(y_0-\frac{pq}{D}\right)^2.
\label{eq:proofbench-advanced-distances}
\end{align}
First observe that
\begin{equation}
1-\frac{p^2}{D}=\frac{D-p^2}{D}=\frac{q^2}{D}.
\label{eq:proofbench-advanced-simplification}
\end{equation}
Consider the difference
\begin{align}
\Delta
&=OA^2-OF^2\notag\\
&=\left[(1-p)^2-\left(1-\frac{p^2}{D}\right)^2\right]
 +\left[(y_0-q)^2-\left(y_0-\frac{pq}{D}\right)^2\right].
\label{eq:proofbench-advanced-delta}
\end{align}
Factoring the first bracket as a difference of squares gives
\begin{align}
&(1-p)^2-\left(1-\frac{p^2}{D}\right)^2\notag\\
&=\left(\frac{p^2}{D}-p\right)
  \left(2-p-\frac{p^2}{D}\right)\notag\\
&=-\frac{p(D-p)}{D}
  \left(2-p-\frac{p^2}{D}\right).
\label{eq:proofbench-advanced-first-bracket}
\end{align}
Similarly,
\begin{align}
&(y_0-q)^2-\left(y_0-\frac{pq}{D}\right)^2\notag\\
&=\left(\frac{pq}{D}-q\right)
  \left(2y_0-q-\frac{pq}{D}\right)\notag\\
&=-\frac{q(D-p)}{D}
  \left(2y_0-q-\frac{pq}{D}\right).
\label{eq:proofbench-advanced-second-bracket}
\end{align}
Substituting equations~(\ref{eq:proofbench-advanced-first-bracket}) and (\ref{eq:proofbench-advanced-second-bracket}) into equation~(\ref{eq:proofbench-advanced-delta}), we obtain
\begin{align}
\Delta
=-\frac{D-p}{D}\Bigg[
&p\left(2-p-\frac{p^2}{D}\right)\notag\\[-2pt]
&+q\left(2y_0-q-\frac{pq}{D}\right)
\Bigg].
\label{eq:proofbench-advanced-factored-delta}
\end{align}
Using equation~(\ref{eq:proofbench-advanced-y0}),
\begin{align}
q\left(2y_0-q-\frac{pq}{D}\right)
&=q\left(\frac{D-p}{q}-q-\frac{pq}{D}\right)\notag\\
&=(D-p)-q^2-\frac{pq^2}{D}.
\label{eq:proofbench-advanced-y0-substitution}
\end{align}
Consequently, the bracket in equation~(\ref{eq:proofbench-advanced-factored-delta}) equals
\begin{align*}
&p\left(2-p-\frac{p^2}{D}\right)
 +(D-p)-q^2-\frac{pq^2}{D}\\
&=2p-p^2-\frac{p^3}{D}+D-p-q^2-\frac{pq^2}{D}\\
&=p+\left(D-p^2-q^2\right)-\frac{p(p^2+q^2)}{D}\\
&=p+0-p=0.
\end{align*}
It follows that $\Delta=0$, and hence $OA^2=OF^2$. Therefore, $O$ lies on the perpendicular bisector of $AF$ as well as that of $AP$, so $O$ is the circumcenter of $\triangle AFP$.

\proofbenchstep{Step 5. Bisecting the chord $XY$.}
The circumcenter $O=(1,y_0)$ and the point $C=(1,0)$ have the same $x$-coordinate. Since $BC$ is the $x$-axis, the line $OC$ is perpendicular to $BC$. The chord $XY$ lies on $BC$, and the perpendicular from the center of a circle to a chord bisects that chord. Thus, $C$ is the midpoint of $XY$.
\[
\boxed{C\text{ is the midpoint of }XY.}
\]
\end{proofsolutionbox}
\endgroup

\section{Case Study}
\label{app:case}

This appendix presents two representative instances of direct OPD when the teacher model SU-01 is distilled to Qwen3.5-35B-A3B. In both cases, the student degenerates into an endless repetition—either a
fixed multi‑sentence self‑check block or a single‑token loop. The rollouts
terminate only when hitting the 160k‑token generation limit.

\subsection{Case A: correct answer, then 972 exact self-check repeats (AIME25)}
\label{case:loop-aime25}

The model solves an AIME 2025 divisibility problem, reaches the correct
numerical answer $\boxed{279}$, but then appends the same ten‑sentence
“verification” unit verbatim 972 times. This loop occupies nearly the entire
remaining generation budget and is truncated mid‑unit.

\par\smallskip\noindent\small

\begin{casebox}{Question}
Determine the difference between $N$ and $2025$, where $N$ is the number of
eight-digit positive integers using each of the digits $1,2,\dots,8$ exactly
once that are divisible by $22$.
\end{casebox}

\begin{proofsolutionbox}{Model Output (excerpt)}
\small

\proofbenchstep{Beginning of output.}
We need to solve: there are $8! = 40320$ eight-digit positive integers that use
each of the digits $1,2,3,4,5,6,7,8$ exactly once. Let $N$ be the number of
these integers that are divisible by $22$. Find the difference between $N$ and
$2025$. So we need to compute $N$, then compute $N-2025$. Since
$22 = 2\cdot 11$, we need divisibility by $2$ and by $11$.

\proofbenchstep{At the failure onset (response chars 75{,}000--77{,}150).}
Consequently, $N = 144 \times 16 = 2304$. Finally, $N - 2025 = 2304 - 2025 =
279$. Thus the answer is $\boxed{279}$. I'll also ensure that the solution is
not missing any step.

\proofbenchstep{Repeating pattern --- exact 10-sentence self-check unit (542 characters), repeated 972 times.}
\begin{quote}\small
``I'll also ensure that the solution is not missing any step.''\\
``I'll also ensure that the solution is clear.''\\
``I'll also ensure that the solution is correct.''\\
``I'll also ensure that the solution is comprehensive.''\\
``I'll also ensure that the solution is easy to follow.''\\
``I'll also ensure that the solution is in the required format.''\\
``I'll also ensure that the solution is not too long.''\\
``I'll also ensure that the solution is self-contained.''\\
``I'll also ensure that the solution is not missing any justification.''\\
``I'll also ensure that the solution is correct.''
\end{quote}
\noindent\small
The model reaches the correct answer $279$, then repeats the same ten-sentence
self-check unit exactly 972 times. The rollout hits the 160k-token
generation limit inside this unit and is recorded as truncated; the response
ends mid-unit with ``\dots\ not missing any justification. I'll also ensure
that the''.
\end{proofsolutionbox}

\subsection{Case B: reasoning collapses into a one-token loop (AnswerBench)}
\label{case:loop-answerbench}

In this case, the model engages with a complex number‑theoretic definition (“exotic
integers”). After some initial reasoning, it gets stuck on a sub‑problem
involving multiplicative orders and begins emitting the token sequence
``$2^?$'' indefinitely. The repetition is atomic (one token) and repeats over
34,000 times with no progress or termination.

\par\smallskip\noindent\small

\begin{casebox}{Question}
For a positive integer $n$, call $g\colon \mathbb{Z}\to\mathbb{Z}$ an
$n$-good function if $g(1)=1$ and, for distinct integers $a$ and $b$,
$g(a)-g(b)$ divides $a^n-b^n$. Call $n$ exotic if the number of $n$-good
functions is twice an odd integer. Find the 132nd exotic integer.
\end{casebox}

\begin{proofsolutionbox}{Model Output (excerpt)}
\small

\proofbenchstep{Beginning of output.}
We need to solve: for positive integer $n$, define $g\colon \mathbb{Z}\to
\mathbb{Z}$ such that $g(1)=1$ and for any two distinct integers $a$ and $b$,
$g(a)-g(b)$ divides $a^n-b^n$. We call $n$ an exotic integer if the number of
$n$-good functions is twice an odd integer. Find the 132nd exotic integer. We
need to understand the condition and count the number of such functions for a
given $n$.

\proofbenchstep{At the failure onset (response tokens 840--919).}
For $m=8$: $2^8-1=255$, order of $2$ mod $255$? $2^8 = 256 \equiv 1 \pmod{255}$,
but $2^4 = 16 \pmod{255}$, not $1$; \dots\ but could be smaller? $2^?$ Actually
\dots\ $2^?$ $2^?$ $2^?$

\proofbenchstep{Repeating pattern --- exact one-token unit, repeated many times.}
\[
2^? \;\; 2^? \;\; 2^? \;\; 2^? \;\; 2^? \;\; 2^? \;\; 2^? \;\; 2^? \;\;
2^? \;\; 2^? \quad\cdots\quad
\]
\noindent\small
The identical token occupies one uninterrupted run; no new claim, calculation, or answer is produced after the loop
begins. The rollout reaches the 160k-token generation limit inside this
run and is recorded as truncated; the response ends with
``\dots\ $2^?$ $2^?$ $2^?$ $2^?$ $2^?$''.
\end{proofsolutionbox}

\end{document}